\documentclass[sigconf,screen]{acmart}

\AtBeginDocument{
  }

\usepackage[T1]{fontenc}
\usepackage[utf8]{inputenc}
\usepackage{textcomp}
\usepackage{amsmath}
\usepackage{amsfonts}
\usepackage{booktabs}
\usepackage[table]{xcolor}
\usepackage{array}
\usepackage{makecell}
\usepackage{multirow}
\usepackage{adjustbox}
\usepackage{arydshln}
\usepackage{graphicx}
\usepackage{hyperref}
\usepackage{pifont}
\usepackage{fontawesome5}
\usepackage[skins]{tcolorbox}

\renewcommand\footnotetextcopyrightpermission[1]{}
\begin{document}

\title{Holtercare-Bench: A Multimodal Benchmark for Evaluating Long-Term Dynamic ECG Analysis}

\author{
Yihan Xie\textsuperscript{1}\footnotemark[1] \quad
Hanwen Cui\textsuperscript{2}\footnotemark[1] \quad
Runze Ye\textsuperscript{1}\footnotemark[1] \quad
Juekai Lin\textsuperscript{1} \quad
Haoyang Wang\textsuperscript{1} \quad
Jinhao Mao\textsuperscript{1} \\
Bo Zhang\textsuperscript{3} \quad
Wenqiao Zhang\textsuperscript{1}\footnotemark[2] \quad
Xiaogang Guo\textsuperscript{1}\footnotemark[2] \quad
Jun Xiao\textsuperscript{1}\footnotemark[2] \quad
Lei Zhang\textsuperscript{1}
}
\email{\{yihanxie, wenqiaozhang\}@zju.edu.cn}
\affiliation{
  \institution{
  \textsuperscript{1}Zhejiang University \quad
  \textsuperscript{2}Beijing Institute of Technology \quad
  \textsuperscript{3}University of Electronic Science and Technology of China
  % \textsuperscript{4}The First Affiliated Hospital, Zhejiang University School of Medicine
  }
  \country{}
}

\begin{abstract}

\begingroup
\renewcommand\thefootnote{}
\footnotetext{$^\ast$These authors contributed equally to this research.}
\footnotetext{$^\dagger$Corresponding authors.}
\endgroup

While multimodal large language models (MLLMs) excel in medical applications, most of them favor static images or short-term signals. In the critical field of dynamic electrocardiograms (ECG), models struggle with complex temporal reasoning and diagnostic report generation due to a lack of high-quality datasets and benchmarks. To address this, we introduce \textbf{(i) Holtercare-23K}, a large-scale multimodal dynamic ECG dataset comprising 22,980 QA pairs derived from 788 clinical Holter records and featuring a novel signal--video--text tri-modal alignment. Based on this dataset, we present \textbf{(ii) Holtercare-Bench}, a multimodal benchmark that evaluates models on temporal localization, clinical diagnosis, and global summarization. Zero-shot evaluations of leading MLLMs reveal a significant performance gap in processing ultra-long pathological sequences. However, fine-tuning representative models yields substantial improvements. This work illuminates the limitations of current MLLMs in electrophysiology and provides a foundational benchmark for long-term medical MLLMs. Our project is available at \url{https://github.com/ZJU4HealthCare/Holtercare-Bench}.

\end{abstract}

\begin{teaserfigure}
\centering
\includegraphics[width=0.98\textwidth]{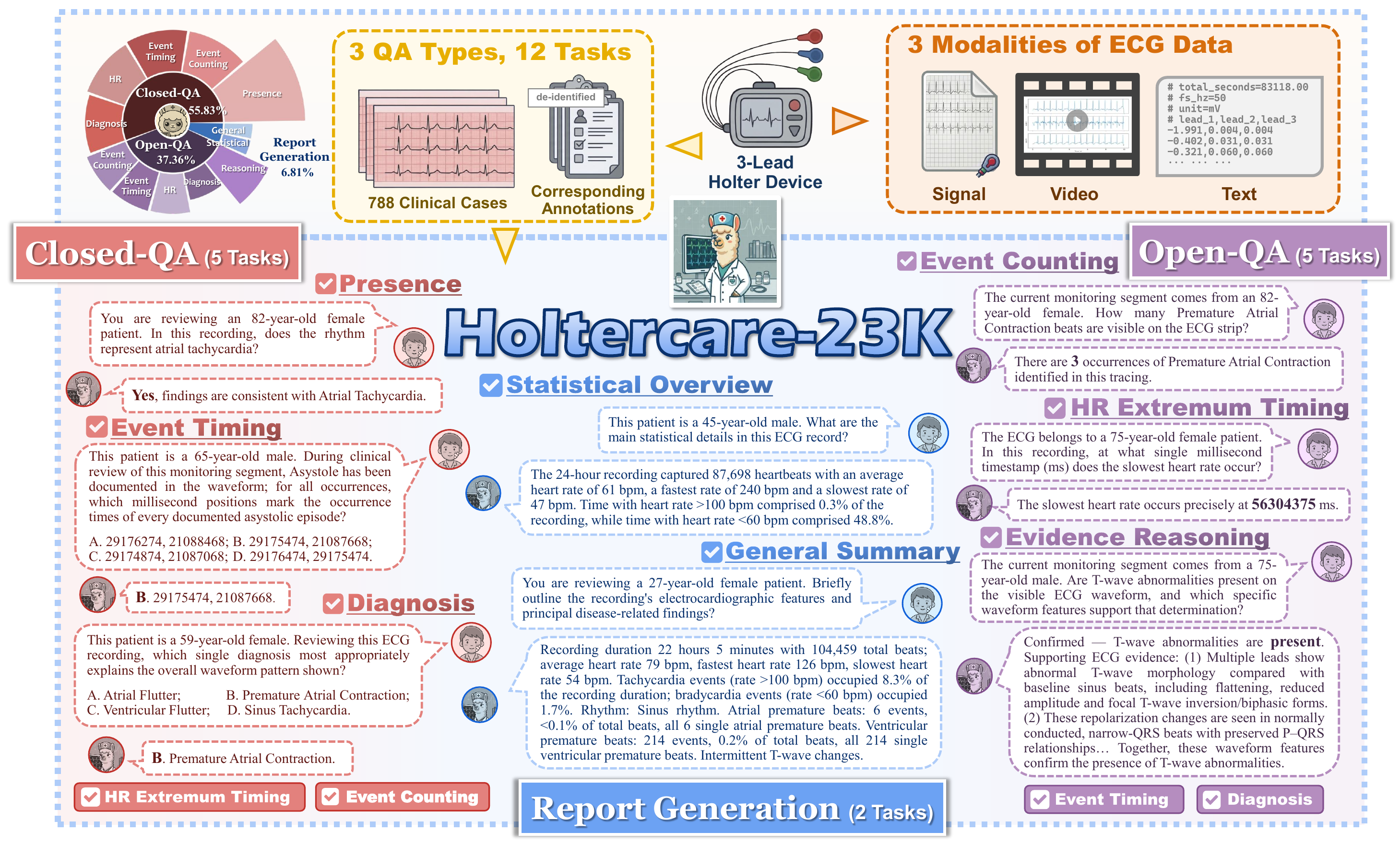}
% \vspace{-3mm}
\caption{Overview of our proposed Holtercare-23K dataset.}
% \vspace{-3mm}
\label{fig:teaser}
\end{teaserfigure}

\maketitle
\pagestyle{empty}

\section{Introduction}

Recent advances in multimodal large language models (MLLMs) have revolutionized medical artificial intelligence (AI). However, current research is heavily skewed toward spatial modalities, such as radiological images and pathology slides. In cardiology, while short-term electrocardiogram (ECG) datasets like PTB-XL~\cite{PhysioNet-ptb-xl-1.0.3} and MIMIC-IV-ECG~\cite{PhysioNet-mimic-iv-ecg-1.0} have driven crucial progress, they only capture a few seconds of cardiac activity. In clinical practice, continuous dynamic ECG Holter monitoring is the gold standard for diagnosing intermittent arrhythmias. Unfortunately, the underdevelopment of the data ecosystem for Holter monitoring has left modern MLLMs untested and unoptimized for long-term cardiac care.

Analyzing dynamic ECGs presents unique challenges that cannot be addressed by short-segment datasets or benchmarks. A standard Holter record spans around 24 hours, containing hundreds of thousands of heartbeats. Within this massive temporal context, critical pathological events---such as a brief episode of ventricular tachycardia---are extremely fleeting. Finding them requires processing an ultra-long sequence while maintaining temporal sensitivity at the millisecond (ms) level. Furthermore, clinical diagnosis is not just about classification; it requires linking micro-level waveform changes (e.g., missing P-waves) and macro-level disease labels. Current MLLMs lack both high-quality data and specific benchmarks for learning this complex clinical reasoning.

To bridge this gap, we introduce \textbf{Holtercare-Bench}, a multimodal benchmark for evaluating long-term dynamic ECG analysis. The main contributions of our work are as follows:

\noindent\textbf{(i) Dataset.} We propose \textbf{Holtercare-23K}, a large-scale dynamic ECG dataset containing 788 real-world clinical records, lasting 13--24 hours each. A major hurdle in applying MLLMs to electrophysiology is the modality mismatch: modern language models cannot naturally process raw voltage arrays. To overcome this, we design \textbf{HolterAgent}, an automated data engine to convert Holter records into a \textbf{signal--video--text} tri-modal format. Additionally, we construct a total of 22,980 QA pairs across three distinct types (\textit{Closed-QA}, \textit{Open-QA}, and \textit{Report Generation}) from expert clinical annotations and reports, bridging the gap between electrophysiological signals and modern MLLM architectures. An overview of the complete Holtercare-23K dataset is illustrated in Figure~\ref{fig:teaser}.

\noindent \textbf{(ii) Benchmark.} Based on Holtercare-23K, we propose \textbf{Holtercare-Bench}, a comprehensive evaluation framework designed to quantify a model's cognitive ability in long-context cardiology. Holtercare-Bench consists of 12 fine-grained tasks that mirror the actual diagnostic workflow of a cardiologist. Instead of relying on a single generic metric, we deploy a tailored scoring system. This ranges from matching accuracy for discrete decision-making to an LLM-as-a-judge system for evaluating the logical consistency of generated clinical reports.

We evaluate a wide range of mainstream generalist and medical MLLMs on our benchmark. Zero-shot results expose a significant performance gap---most current models struggle to maintain temporal consistency over ultra-long sequences. However, fine-tuning on our dataset leads to substantial performance gains, demonstrating the efficacy of Holtercare-23K in enhancing complex clinical reasoning. Our results establish a strong baseline and define the performance frontiers for future long-context medical MLLMs.
\section{Related Work}

\textbf{Medical MLLMs and ECG Representation Learning.} Recent MLLMs have progressively advanced from general vision--language understanding toward fine-grained spatial--temporal perception, instruction-driven visual reasoning, and adaptive multimodal architectures. Representative efforts include VideoRefer~\cite{yuan2025videorefer} and PixelRefer~\cite{yuan2025pixelrefer} for spatial-temporal object understanding, InstructSAM~\cite{yuan2026instructsam} for instruction-driven visual segmentation, VisualThink-VLA~\cite{gao2026visualthink} for visual intermediate reasoning, and HyperLLaVA~\cite{zhang2024hyperllava} for dynamic visual--language adaptation. These advances have also extended to the medical domain, where MLLMs such as Med-Flamingo~\cite{moor2023med}, LLaVA-Med~\cite{li2023llava}, MedVLM~\cite{pan2025medvlm}, Lingshu~\cite{xu2025lingshu}, and HealthGPT~\cite{lin2025healthgpt} have advanced multimodal clinical understanding. Specialized MLLMs have further emerged across diverse clinical domains, including LLaVA-Rad~\cite{chaves2024towards} in radiology, EyecareGPT~\cite{li2025eyecaregpt} in ophthalmology, and SkinGPT~\cite{zhou2023skingpt} in dermatology. More recent works have moved toward modality-specific modeling and clinically grounded reasoning, including unified slice-volume analysis in OmniCT~\cite{lin2026omnict}, multimodal chain-of-thought (CoT) reasoning in TumorChain~\cite{li2026tumorchain}, Group Relative Policy Optimization (GRPO) in TIF-GRPO~\cite{lin2026regulating}, and evidence-driven multimodal reinforcement learning in E-MRL~\cite{li2026emrl}.

In the ECG domain, models like ECGFounder~\cite{li2025electrocardiogram} and ECG-FM~\cite{mckeen2025ecg} enhance pretraining through massive datasets and a combination of self-supervised learning techniques. Multimodal integration is also advancing: ECG-SL~\cite{yu2023ecg}, HeartLang~\cite{jin2025reading}, and ESI~\cite{yu2024ecg} focus on signal-semantic alignment, whereas ECG-LM~\cite{yang2025ecg}, SuPreME~\cite{cai2025supreme}, MERL~\cite{liu2024zero}, and KED~\cite{tian2024foundation} leverage external clinical knowledge to improve representations. Concurrently, generative paradigms bridge LLMs and ECGs through continuous voltage tokenization as in ECG-Byte~\cite{han2024ecg}, instruction tuning for report generation like MEIT~\cite{wan2025meit}, or visual adaptation as in PULSE~\cite{liu2024teach}. Despite these advances, current research generally addresses isolated tasks or short-duration static inputs, lacking a unified framework for complex reasoning over ultra-long Holter contexts.

\begin{table}[ht]
\centering
% \small
% \renewcommand{\arraystretch}{1.2}
% \vspace{-3mm}
\caption{Systematic comparison of current ECG datasets.}
% \vspace{-3mm}
\label{tab:dataset_comparison}
\begin{adjustbox}{width=0.98\linewidth,center}
\begin{tabular}{lcccccc}
\toprule
\multirow{2}{*}{\textbf{Dataset}} & \multirow{2}{*}{\textbf{Leads}} & \multirow{2}{*}{\textbf{Duration}} & \multirow{2}{*}{\textbf{Size}} & \multicolumn{3}{c}{\textbf{Annotation}} \\
\cmidrule(lr){5-7}
& & & & \textbf{Beat} & \textbf{Rhythm} & \textbf{Report} \\

\hline
\rowcolor{gray!10}
\multicolumn{7}{c}{\textbf{\textit{Static ECG Datasets}}} \\
\hline
\rowcolor{cyan!3}
PTB-XL~\cite{PhysioNet-ptb-xl-1.0.3} & 12 & 10 seconds & 21,837 & & & \checkmark \\
\rowcolor{cyan!3}
ecg-arrhythmia~\cite{PhysioNet-ecg-arrhythmia-1.0.0} & 12 & 10 seconds & 45,152 & & \checkmark & \\
\rowcolor{cyan!3}
MIMIC-IV-ECG~\cite{PhysioNet-mimic-iv-ecg-1.0} & 12 & 10 seconds & 800K & & & \checkmark \\
\rowcolor{cyan!3}
CPSC 2018~\cite{liu2018open} & 12 & 6--60 seconds & 9,831 & & \checkmark & \\
\rowcolor{cyan!3}
LUDB~\cite{PhysioNet-ludb-1.0.1} & 12 & 10 seconds & 200 & \checkmark & & \\

\hline
\rowcolor{gray!10}
\multicolumn{7}{c}{\textbf{\textit{Dynamic ECG Datasets}}} \\
\hline
\rowcolor{lime!3}
MITDB~\cite{moody2001impact} & 2 & 30 minutes & 48 & \checkmark & & \\
\rowcolor{lime!3}
INCARTDB~\cite{incartdb} & 12 & 30 minutes & 75 & \checkmark & & \\
\rowcolor{lime!3}
LTSTDB~\cite{jager2003long} & 2--3 & 21--24 hours & 86 & \checkmark & \checkmark & \\
\rowcolor{lime!3}
Icentia11k~\cite{PhysioNet-icentia11k-continuous-ecg-1.0} & 1 & 70 minutes & 11K & \checkmark & \checkmark & \\
\rowcolor{lime!3}
AFDB~\cite{moody1983new} & 2 & 10 hours & 25 & \checkmark & \checkmark & \\
\rowcolor{lime!3}
LTDB~\cite{ltdb} & 2 & 14--22 hours & 7 & \checkmark & & \\
\rowcolor{lime!3}
SDDB~\cite{greenwald1986development} & 2 & 14--25 hours & 23 & \checkmark & & \\
\rowcolor{lime!3}
NSRDB~\cite{nsrdb} & 2 & 23--26 hours & 18 & \checkmark & & \\
\rowcolor{lime!3}
LTAFDB~\cite{petrutiu2007abrupt} & 2 & 24--25 hours & 84 & \checkmark & \checkmark & \\
\rowcolor{lime!3}
SHDB-AF~\cite{PhysioNet-shdb-af-1.0.1} & 2 & 24 hours & 143 & \checkmark & & \\
\rowcolor{lime!3}
QTDB~\cite{laguna1997database} & 2 & 15 minutes & 105 & \checkmark & & \\
\rowcolor{lime!3}
Apnea-ECG~\cite{penzel2000apnea} & 1 & 7--10 hours & 70 & \checkmark & \checkmark & \\
\rowcolor{orange!20}
\textbf{Holtercare-23K (Ours)} & \textbf{3} & \textbf{13--24 hours} & \textbf{788} & \checkmark & \checkmark & \checkmark \\

\bottomrule
\end{tabular}
\end{adjustbox}
% \vspace{-3mm}
\end{table}

\noindent \textbf{ECG Analysis Datasets.} Public ECG datasets like PTB-XL~\cite{PhysioNet-ptb-xl-1.0.3} and MIMIC-IV-ECG~\cite{PhysioNet-mimic-iv-ecg-1.0} have driven deep learning diagnostics but consist of short signal segments. Consequently, they fail to capture long-term arrhythmia dynamics. For continuous monitoring, dynamic datasets like MITDB~\cite{moody2001impact}, LTSTDB~\cite{jager2003long}, and LTAFDB~\cite{petrutiu2007abrupt} have been introduced. However, the critical limitations of these are shown in Table~\ref{tab:dataset_comparison}. They are either too small to train large-scale models or feature limited annotations confined to simple discrete labels and beat-level localizations. They severely lack fine-grained descriptions required for complex cross-modal analysis and clinical report generation. This gap prevents models from learning causal clinical reasoning and underscores the critical need for large-scale data with multimodal alignments.

\noindent \textbf{Medical Benchmarks.} Current benchmarks are inadequate for long-term ECG analysis. Recent large-scale medical VQA benchmarks (e.g., PMC-VQA~\cite{zhang2023pmc}, GMAI-MMBench~\cite{chen2024gmai}) offer thorough evaluation frameworks across various clinical modalities; however, along with previous datasets such as VQA-RAD~\cite{lau2018dataset} and Slake~\cite{liu2021slake}, they are strictly limited to static spatial images. Even within the cardiovascular field, current benchmarks like ECG-QA~\cite{oh2023ecg} primarily emphasize short-duration static waveforms rather than continuous monitoring. In contrast, general long-sequence benchmarks (Video-MME~\cite{fu2025video}, LVBench~\cite{wang2025lvbench}) focus on ordinary videos lacking specialized medical reasoning. Evaluating dynamic ECGs demands high-dimensional pathological feature extraction and temporal dependency modeling, yet the community currently lacks a comprehensive benchmark to systematically evaluate MLLMs on continuous, long-term clinical reasoning.
\section{Dataset: Holtercare-23K}

To address the scarcity of long-term dynamic ECG data for MLLM research, we construct \textbf{Holtercare-23K}, a large-scale, multimodal dynamic ECG dataset derived from 788 clinically collected dynamic ECGs, which contains 22,980 QA pairs divided into three categories.

\subsection{Dataset Overview}

\begin{figure*}[ht]
\includegraphics[width=0.98\linewidth]{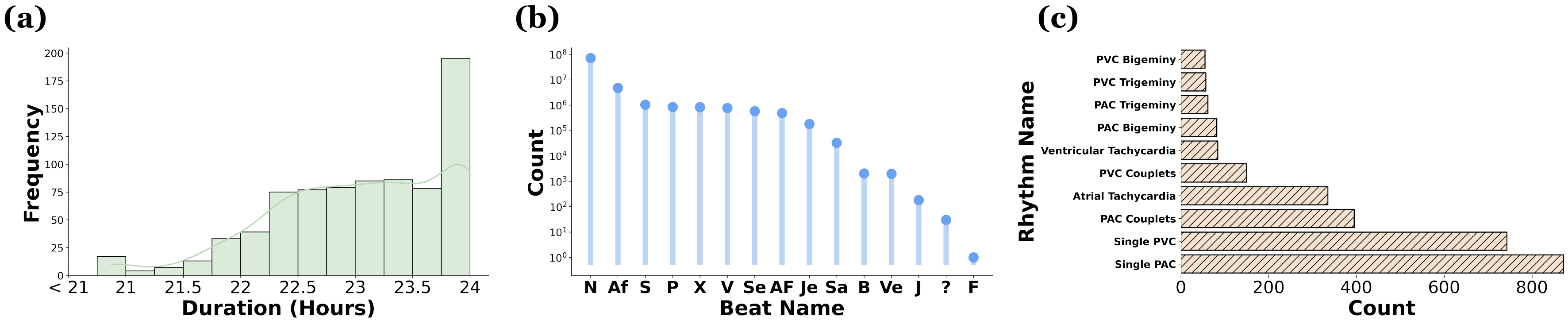}
% \vspace{-3mm}
\caption{Statistical distribution of Holtercare-23K. (a) Density estimation of recording durations. (b) Log-scale counts of beat-level annotations. (c) Frequencies of the top 10 rhythm-level events.}
\label{fig:count}
% \vspace{-3mm}
\end{figure*}

All data in Holtercare-23K originates from real-world hospital collections recorded in 2026. Compared with existing public databases in Table~\ref{tab:dataset_comparison}, the dataset comprises 788 independent cases, with each continuous monitoring session lasting between 13 and 24 hours, as detailed in Figure~\ref{fig:count} (a). To ensure profound medical utility, every record is equipped with a comprehensive, tri-level annotation system:

\noindent \textbf{(i) Beat-Level Annotations.} As Figure~\ref{fig:count} (b) shows, these annotations pinpoint the exact occurrence timestamps for 18 distinct heartbeat categories (e.g., normal beats, premature contractions) and various non-beat events (e.g., P-waves, T-waves, artifacts).

\noindent \textbf{(ii) Rhythm-Level Annotations.} These annotations capture continuous cardiac events and arrhythmias and provide specific string descriptions (e.g., ventricular tachycardia, ST-segment changes) bound by precise timestamps, with the top 10 frequent events shown in Figure~\ref{fig:count} (c).

\noindent \textbf{(iii) Report-Level Summaries.} Each case includes a global report strictly verified by professional cardiologists. These reports provide overall recording statistics (e.g., total heartbeats, average heart rate, exact timestamps for maximum/minimum heart rates, and duration percentages for tachycardia/bradycardia) along with final diagnostic conclusions.

Regarding data privacy and compliance, all raw data underwent rigorous de-identification. We remove all sensitive personal identifiers but preserve critical medical context: age, gender, and anonymize electronic medical records (EMRs). These EMRs provide brief clinical histories and primary symptoms (e.g., ``history of hypertension,'' ``dizziness,'' or ``chest tightness''), serving as vital supplementary inputs for multimodal clinical reasoning. The entire dataset construction fully complies with medical ethics and data protection laws.

\subsection{Multimodal Data Engine: HolterAgent}

To transform this massive, multi-grained clinical data into QA pairs suitable for MLLM evaluation, we develop an automated multimodal data engine, \textbf{HolterAgent}. The systematic workflow of HolterAgent, from signal preprocessing to QA generation, is illustrated in Figure~\ref{fig:engine}. The processing pipeline involves three core modules:

\begin{figure}[ht]
% \vspace{-3mm}
\includegraphics[width=0.98\linewidth]{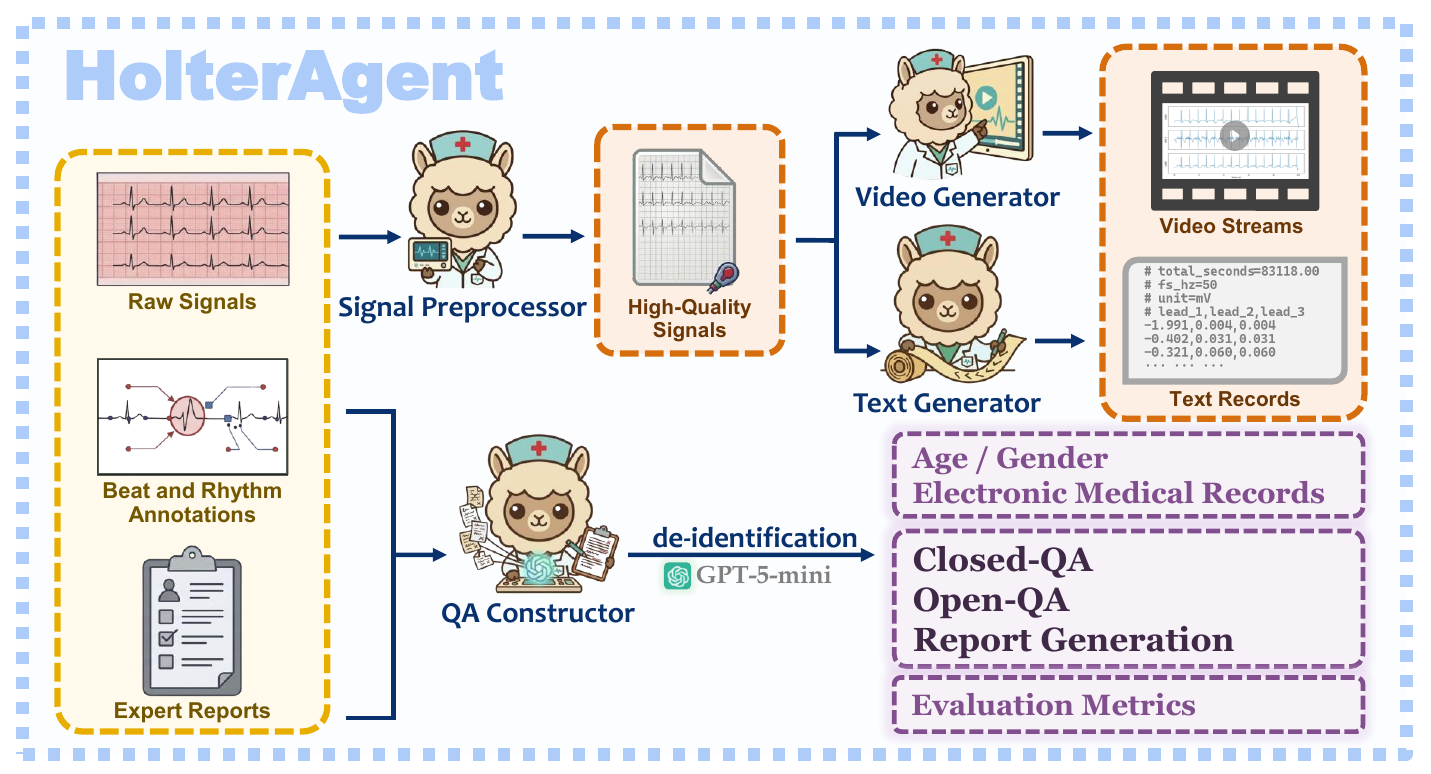}
% \vspace{-3mm}
\caption{The multimodal generation and data construction framework of HolterAgent.}
\label{fig:engine}
% \vspace{-5mm}
\end{figure}

\noindent \textbf{(i) Signal Preprocessor.} To initiate the pipeline, HolterAgent first employs the \texttt{MNE}~\cite{gramfort2014mne} package to parse the raw clinical EDF files and extract continuous electrophysiological signals. Considering the inevitable motion artifacts and baseline wandering in real clinical environments, we subsequently utilize the \texttt{NeuroKit2}~\cite{makowski2021neurokit2} package to perform lead-level deep denoising and baseline correction. This cascaded process ensures the extraction of the purest pathological waveform features for downstream tasks.

\noindent \textbf{(ii) Video and Text Generator.} To accommodate diverse MLLM architectures, HolterAgent transforms these cleaned, high-quality signals into two extra modalities: clinical text records and dynamic video streams. For text modality, the engine extracts the specified lead data, scales it to millivolts (mV), and formats it to three decimal places. For video modality, the engine employs \texttt{Matplotlib}~\cite{hunter2007matplotlib} to construct sliding windows (e.g., a 10-second duration), rendering the long-term signals into dynamic ECG video streams.

\noindent \textbf{(iii) QA Constructor.} Relying on the preprocessed signals and the tri-level annotations, we employ GPT-5-mini~\cite{gpt5} to perform deep semantic parsing and logical restructuring. HolterAgent automatically transforms this rich clinical context into various QA types, yielding a total of 22,980 high-quality, multimodal QA pairs.

To ensure rigorous model evaluation and strictly prevent data leakage, we partition the dataset at the independent case level (788 cases in total) rather than the QA pair level. Specifically, we randomly hold out 20\% of the cases to form the test set. The remaining 80\% of the cases are further divided into training and validation sets following a 9:1 ratio. Consequently, the 22,980 QA pairs are seamlessly distributed into their respective splits based on their source cases, guaranteeing that no patient data overlaps between the training, validation, and test phases. An illustrative overview of Holtercare-23K is depicted in Figure~\ref{fig:teaser}.
\section{Benchmark: Holtercare-Bench}

To systematically evaluate the cognitive and multimodal reasoning capabilities of MLLMs on long-term dynamic ECGs, we design \textbf{Holtercare-Bench} to mirror the diagnostic workflow of human cardiologists. The benchmark is structured into three progressive evaluation tiers: \textit{Closed-QA}, \textit{Open-QA}, and \textit{Report Generation}. This yields a total of 12 fine-grained multimodal tasks, each equipped with rigorous, objective evaluation metrics.

\subsection{Closed-QA Tasks}

\textit{Closed-QA} tasks evaluate the fundamental feature recognition and discrete decision-making abilities of models. By providing explicit multiple-choice candidates, these tasks assess absolute precision using strict matching accuracy. This module includes five sub-tasks:

\noindent \textbf{(i) \textit{Presence}.} Requires the model to determine whether a specific rhythm or abnormal waveform exists within the complex sequence.

\noindent \textbf{(ii) \textit{Event Counting}.} Requires the model to select the correct occurrence count for a specific anomaly category from multiple options.

\noindent \textbf{(iii) \textit{Event Timing}.} Requires the model to select the precise ms-level timestamp combination for all occurrences of an abnormal event from highly similar interfering candidates.

\noindent \textbf{(iv) \textit{HR Extremum Timing}.} Requires the model to accurately match the ms-level timestamp of the fastest or slowest heart rate (HR) amidst global rhythm variations.

\noindent \textbf{(v) \textit{Diagnosis}.} Requires the model to select the comprehensive diagnosis that best matches the overall characteristics of the current long-term strip from multiple similar arrhythmia category options.

\subsection{Open-QA Tasks}

\textit{Open-QA} tasks require models to autonomously generate free-text answers. We evaluate text quality using \textit{BLEU}~\cite{papineni2002bleu}, \textit{ROUGE-L}~\cite{lin2004rouge}, and \textit{F1-Bio}~\cite{ramshaw1995text}, along with a custom normalized metric \textit{Score\textsuperscript{MAE}} for numerical reasoning. This module includes five sub-tasks:

\noindent \textbf{(i) \textit{Event Counting} and (ii) \textit{HR Extremum Counting}.} To evaluate numerical accuracy while preventing metric collapse for baselines, we transform the mean absolute error (MAE) into a normalized score: $Score^{MAE} = 100 / [1 + \exp((MAE - \mu)/\sigma)]$, where $\mu$ (median) and $\sigma$ (standard deviation) anchor baseline scores around 50. This non-linear scaling clearly distinguishes fine-tuned models without severely compressing baselines.

\noindent \textbf{(iii) \textit{Event Timing} and (iv) \textit{Diagnosis}.} Models autonomously output structured ms-level temporal offsets or multi-label diagnoses, which are scored for precision and completeness using the text and overlap metrics described above.

\noindent \textbf{(v) \textit{Evidence Reasoning}.} Models must generate rationales (e.g., ``absence of P waves'') to support their diagnoses, demonstrating a causal logic chain from raw waveforms to clinical decisions.

\begin{table}[ht]
\centering
% \small
% \renewcommand{\arraystretch}{1.1}
% \vspace{-3mm}
\caption{Evaluation dimensions and granular criteria for reports generated on \textit{Statistical Overview} tasks.}
\label{tab:metric_stat}
% \vspace{-3mm}
\begin{adjustbox}{width=0.98\linewidth,center}
\begin{tabular}{p{1.7cm}p{6.4cm}c}
\toprule
\textbf{Dimension} & \textbf{Evaluation Criteria} & \textbf{Weight} \\
\midrule

\multirow{2}{1.7cm}{Global Metrics}
& Accurate extraction of recording duration & 15 \\
\cmidrule(l){2-3}
& Accurate extraction of the total heartbeat count & 10 \\
\midrule

\multirow{3}{1.7cm}{Heart Rate Stats}
& Accurate reporting of the average heart rate & 10 \\
\cmidrule(l){2-3}
& Accurate reporting of the maximum heart rate & 10 \\
\cmidrule(l){2-3}
& Accurate reporting of the minimum heart rate & 10 \\
\midrule

\multirow{4}{1.7cm}{Rhythm Burden}
& Correct extraction of tachycardia burden (percentage of time HR > 100 bpm) & 15 \\
\cmidrule(l){2-3}
& Correct extraction of bradycardia burden (percentage of time HR < 60 bpm) & 15 \\
\midrule

\multirow{4}{1.7cm}{Report Logic and Norms}
& Logically organized and coherent natural language & 5 \\
\cmidrule(l){2-3}
& Adherence to clinical reporting conventions & 5 \\
\cmidrule(l){2-3}
& Restriction to statistical metrics without introducing clinical diagnoses & 5 \\
\midrule

\textbf{Total Score} & & \textbf{100} \\
\bottomrule
\end{tabular}
\end{adjustbox}
% \vspace{-3mm}
\end{table}

\subsection{Report Generation Tasks}

\textit{Report Generation} tasks require the model to step beyond local features and perform global abstraction over long-term records, directly mimicking real-world Holter reporting. To ensure professional evaluation, we employ medical text metrics (\textit{BLEU}~\cite{papineni2002bleu}, \textit{ROUGE-L}~\cite{lin2004rouge}, \textit{F1-Bio}~\cite{ramshaw1995text}) and an innovative LLM-as-a-judge framework based on GPT-5-mini~\cite{gpt5}, which yields a comprehensive metric denoted as \textit{Score\textsuperscript{GPT}}. The evaluation rubric and the automated scoring consistency have been cross-verified by professional cardiologists, ensuring clinical accuracy, completeness, and compliance. This module includes two sub-tasks:

\noindent \textbf{(i) \textit{Statistical Overview}.} Requires the model to extract and organize key clinical metrics (e.g., total heartbeat count, extreme heart rates, anomaly burdens) from long-sequence data, ensuring global logical self-consistency. Scoring follows the criteria in Table~\ref{tab:metric_stat}.

\noindent \textbf{(ii)\textit{ General Summary}.} Requires the model produce a summary in the style of a clinical report, identifying key findings by combining local temporal patterns with global statistics. Scoring follows the criteria in Table~\ref{tab:metric_general}.

\begin{table}[ht]
\centering
% \small
% \renewcommand{\arraystretch}{1.1}
% \vspace{-3mm}
\caption{Evaluation dimensions and granular criteria for reports generated on \textit{General Summary} tasks.}
\label{tab:metric_general}
% \vspace{-3mm}
\begin{adjustbox}{width=0.98\linewidth,center}
\begin{tabular}{p{1.6cm}p{8.5cm}c}
\toprule
\textbf{Dimension} & \textbf{Evaluation Criteria} & \textbf{Weight} \\
\midrule

\multirow{3}{1.6cm}{Statistical Accuracy}
& Accurate extraction of global metrics & 4 \\
\cmidrule(l){2-3}
& Accurate reporting of HR extremes & 6 \\
\cmidrule(l){2-3}
& Correct extraction of tachycardia and bradycardia burdens & 4 \\
\midrule

\multirow{3}{1.6cm}{Ectopic Beat and Rhythm Detail}
& Correct classification of the baseline rhythm (e.g., AFib) & 8 \\
\cmidrule(l){2-3}
& Accurate total count and burden of PACs/PVCs & 7 \\
\cmidrule(l){2-3}
& Precise breakdown of ectopic patterns (e.g., isolated, paired, runs) & 10 \\
\midrule

\multirow{3}{1.6cm}{Significant Findings}
& Complete inclusion of severe events (e.g., VT runs, VF, asystole) & 10 \\
\cmidrule(l){2-3}
& Correct interpretation of conduction blocks or pacing signals & 8 \\
\cmidrule(l){2-3}
& Correct reporting of ST-segment and T-wave changes & 7 \\
\midrule

\multirow{3}{1.6cm}{Factual Fidelity}
& Strict consistency with clinical facts without hallucinated findings & 8 \\
\cmidrule(l){2-3}
& Objective description without exaggeration or understatement & 6 \\
\cmidrule(l){2-3}
& Clinical diagnoses strictly grounded in supporting statistical data & 6 \\
\midrule

\multirow{3}{1.6cm}{Structural Logic}
& Coherent narrative structure without fragmented data enumeration & 4 \\
\cmidrule(l){2-3}
& Structured clinical hierarchy, prioritizing major diagnoses over secondary findings & 3 \\
\midrule

\multirow{4}{1.6cm}{Descriptive Norms}
& Precise application of medical terminology & 4 \\
\cmidrule(l){2-3}
& Consistent numeric formatting and standardized unit application (e.g., ``bpm'' for rate, ``\%'' for burden) & 3 \\
\cmidrule(l){2-3}
& Inclusion of appropriate clinical caveats & 2 \\
\midrule

\textbf{Total Score} & & \textbf{100} \\
\bottomrule
\end{tabular}
\end{adjustbox}
% \vspace{-3mm}
\end{table}

\textit{Score\textsuperscript{GPT}} scores both generation task on a 100-point rubric, defined as follows:

\begin{itemize}
\item \textbf{Perfect (90--100):} Perfect or near-perfect compliance with this specific criterion.
\item \textbf{Substantial (70--89):} Substantial compliance, with minor flaws, slight inaccuracies, or trivial omissions.
\item \textbf{Partial (40--69):} Captures the relevant clinical concept but applies an incorrect metric type or statistical aggregation.
\item \textbf{Poor (10--39):} Barely related or severely flawed, but not completely blank.
\item \textbf{Failure (0--9):} Complete failure; the concept is entirely missing or severely hallucinated.
\end{itemize}
\section{Experiments}

To validate Holtercare-Bench and the fine-tuning potential of the Holtercare-23K dataset, we conduct a two-stage experiment: a zero-shot baseline evaluation of mainstream models, followed by a comparative analysis of two representative o
pen-source models before and after fine-tuning.

\begin{table}[ht]
\centering
% \renewcommand{\arraystretch}{1.2}
% \vspace{-3mm}
\caption{Performance comparison of evaluated models on \textit{Closed-QA} tasks from Holtercare-Bench, evaluated by accuracy. \textbf{Bold} and \underline{underline} indicate the best and second-best results, and \faFont\ and \faFilm\ indicate the text and video modalities, respectively. Superscript \textsuperscript{FT} denotes fine-tuned models.}
\label{tab:base_closed}
% \vspace{-3mm}
\begin{adjustbox}{width=0.98\linewidth,center}
\begin{tabular}{lcccccc}
\toprule
\textbf{Model} & \textbf{Modality} & \textbf{Presence} & \makecell{\textbf{Event}\\ \textbf{Counting}} & \makecell{\textbf{Event}\\ \textbf{Timing}} & \makecell{\textbf{HR Extremum}\\ \textbf{Timing}} & \textbf{Diagnosis} \\

\hline
\rowcolor{gray!10}
\multicolumn{7}{c}{\textbf{\textit{Generalist Models}}} \\
\hline
\rowcolor{cyan!3}
GPT-5-mini & \faFont & \underline{76.31} & 31.80 & 33.54 & 38.54 & 48.03 \\
\rowcolor{cyan!3}
Claude-4.5-Haiku & \faFont & 51.63 & 25.99 & 31.71 & 28.34 & 28.35 \\
\rowcolor{cyan!3}
Phi-4-mini-3.8B & \faFont & 52.78 & 27.83 & 33.23 & 36.31 & 24.80 \\
\rowcolor{orange!20}
\textbf{Phi-4-mini-3.8B\textsuperscript{FT}} & \faFont & 64.31 & \underline{53.52} & 63.41 & 54.78 & \underline{49.61} \\
\rowcolor{lime!3}
InternVL-3.5-8B & \faFilm & 67.97 & 14.07 & 40.85 & 26.11 & 42.13 \\
\rowcolor{lime!3}
MiniCPM-V4.5-8B & \faFilm & 52.29 & 19.57 & 37.50 & 39.17 & 19.69 \\
\rowcolor{lime!3}
Gemini-3.0-Flash & \faFilm & 58.17 & 22.63 & \underline{64.33} & \underline{56.37} & 39.76 \\
\rowcolor{lime!3}
Qwen3-VL-8B & \faFilm & 59.15 & 24.46 & 35.98 & 33.12 & 41.73 \\
\rowcolor{orange!20}
\textbf{Qwen3-VL-8B\textsuperscript{FT}} & \faFilm & \textbf{94.77} & \textbf{81.65} & \textbf{99.70} & \textbf{69.11} & \textbf{95.28} \\

\hline
\rowcolor{gray!10}
\multicolumn{7}{c}{\textbf{\textit{Medical Models}}} \\
\hline
\rowcolor{cyan!3}
LLaVA-Med-V1.5-7B & \faFont & 43.79 & 22.32 & 25.91 & 12.10 & 27.95 \\
\rowcolor{cyan!3}
MedGemma-1.5-4B-IT & \faFont & 62.42 & 20.80 & 30.79 & 34.71 & 33.46 \\
\rowcolor{cyan!3}
HealthGPT-M3-3.8B & \faFont & 56.05 & 14.07 & 24.09 & 28.03 & 31.10 \\
\rowcolor{lime!3}
Lingshu-7B & \faFilm & 57.84 & 20.18 & 27.74 & 35.35 & 31.89 \\
\rowcolor{lime!3}
MedVLM-R1-2B & \faFilm & 54.25 & 42.81 & 30.18 & 32.17 & 35.43 \\
\rowcolor{lime!3}
HuatuoGPT-Vision-7B & \faFilm & 50.65 & 18.96 & 25.00 & 36.31 & 21.26 \\

\bottomrule
\end{tabular}
\end{adjustbox}
% \vspace{-3mm}
\end{table}

\subsection{Experimental Setup}

\textbf{Evaluated Models.} We evaluate 13 cutting-edge models, categorized into \textit{Generalist Models} (GPT-5-mini~\cite{gpt5}, Claude-4.5-Haiku~\cite{claude45haiku}, Phi-4-mini-3.8B~\cite{abdin2024phi}, InternVL-3.5-8B~\cite{wang2025internvl3}, MiniCPM-V4.5-8B~\cite{yu2025minicpm}, Gemini-3.0-Flash~\cite{gemini3flash}, and Qwen3-VL-8B~\cite{bai2025qwen3}) and \textit{Medical Models} (LLaVA-Med-V1.5-7B~\cite{li2023llava}, MedGemma-1.5-4B-IT~\cite{sellergren2025medgemma}, HealthGPT-M3-3.8B~\cite{lin2025healthgpt}, Lingshu-7B~\cite{xu2025lingshu}, MedVLM-R1-2B~\cite{pan2025medvlm} and HuatuoGPT-Vision-7B~\cite{chen2024towards}).

\noindent \textbf{Modality Adaptation.} Since these models cannot directly process raw electrophysiological signals, we utilize the tri-modal alignment of Holtercare-23K, evaluating video-capable models (e.g., Qwen3-VL-8B) via video modality and others (e.g., GPT-5-mini) via transformed text modality. Notably, due to inherent model input capacity limits, extended continuous ECG text sequences may be truncated, and video inputs are constrained by file size limits, necessitating accelerated playback or truncation.

\noindent \textbf{Fine-Tuning Setup.} To explore the impact of our dataset on model performance, we select two representative open-source models for instruction tuning: \textbf{Phi-4-mini-3.8B~\cite{abdin2024phi}}, evaluated on text, and \textbf{Qwen3-VL-8B~\cite{bai2025qwen3}}, evaluated on video. We partition Holtercare-23K into training, validation, and test sets to align the models with actual clinical workflows.

\begin{table}[ht]
\centering
\caption{Performance comparison of evaluated models on \textit{Report Generation} tasks from Holtercare-Bench.}
\label{tab:base_report}
% \vspace{-3mm}
\begin{adjustbox}{width=0.98\linewidth,center}
\begin{tabular}{lccccccc}
\toprule
\multirow{2}{*}{\textbf{Model}} & \multirow{2}{*}{\textbf{Modality}} & \multicolumn{3}{c}{\textbf{Statistical Overview}} & \multicolumn{3}{c}{\textbf{General Summary}} \\
\cmidrule(lr){3-5}
\cmidrule(lr){6-8}
& & \textbf{ROUGE-L} $\uparrow$ & \textbf{F1-Bio} $\uparrow$ & \textbf{Score\textsuperscript{GPT}} $\uparrow$ & \textbf{F1-Bio} $\uparrow$ & \textbf{ROUGE-L} $\uparrow$ & \textbf{Score\textsuperscript{GPT}} $\uparrow$ \\

\hline
\rowcolor{gray!10}
\multicolumn{8}{c}{\textbf{\textit{Generalist Models}}} \\
\hline
\rowcolor{cyan!3}
GPT-5-mini & \faFont & 13.68 & 80.56 & 31.08 & 11.03 & 81.39 & 27.28 \\
\rowcolor{cyan!3}
Claude-4.5-Haiku & \faFont & 9.08 & 73.03 & 28.32 & 7.74 & 73.55 & 28.89 \\
\rowcolor{cyan!3}
Phi-4-mini-3.8B & \faFont & 25.92 & 82.26 & 31.63 & 11.51 & 65.57 & 26.77 \\
\rowcolor{orange!20}
\textbf{Phi-4-mini-3.8B\textsuperscript{FT}} & \faFont & \underline{49.94} & \underline{90.63} & \textbf{45.88} & \underline{34.60} & \underline{88.68} & 29.95 \\
\rowcolor{lime!3}
InternVL-3.5-8B & \faFilm & 31.15 & 83.54 & 19.58 & 10.58 & 61.18 & 30.22 \\
\rowcolor{lime!3}
MiniCPM-V4.5-8B & \faFilm & 14.97 & 75.38 & 15.81 & 9.45 & 75.41 & 26.31 \\
\rowcolor{lime!3}
Gemini-3.0-Flash & \faFilm & 16.56 & 71.96 & 22.59 & 10.97 & 75.77 & 28.51 \\
\rowcolor{lime!3}
Qwen3-VL-8B & \faFilm & 13.30 & 67.52 & 15.43 & 10.92 & 69.84 & 24.64 \\
\rowcolor{orange!20}
\textbf{Qwen3-VL-8B\textsuperscript{FT}} & \faFilm & \textbf{51.31} & \textbf{93.23} & \underline{41.03} & \textbf{57.16} & \textbf{95.25} & \textbf{40.79} \\

\hline
\rowcolor{gray!10}
\multicolumn{8}{c}{\textbf{\textit{Medical Models}}} \\
\hline
\rowcolor{cyan!3}
LLaVA-Med-V1.5-7B & \faFont & 15.32 & 74.83 & 13.46 & 8.05 & 62.70 & 16.34 \\
\rowcolor{cyan!3}
MedGemma-1.5-4B-IT & \faFont & 16.09 & 76.84 & 29.93 & 7.94 & 66.54 & \underline{36.48} \\
\rowcolor{cyan!3}
HealthGPT-M3-3.8B & \faFont & 12.81 & 69.85 & 19.27 & 11.49 & 63.98 & 32.26 \\
\rowcolor{lime!3}
Lingshu-7B & \faFilm & 13.36 & 63.59 & 14.73 & 10.45 & 61.51 & 31.37 \\
\rowcolor{lime!3}
MedVLM-R1-2B & \faFilm & 22.97 & 69.92 & 17.29 & 11.99 & 67.91 & 25.90 \\
\rowcolor{lime!3}
HuatuoGPT-Vision-7B & \faFilm & 12.73 & 74.11 & 25.24 & 9.44 & 70.37 & 27.66 \\

\bottomrule
\end{tabular}
\end{adjustbox}
% \vspace{-5mm}
\end{table}
\begin{table*}[ht]
\centering
\caption{Performance comparison of evaluated models on \textit{Open-QA} tasks from Holtercare-Bench.}
\label{tab:base_open}
% \vspace{-3mm}
\begin{adjustbox}{width=0.98\linewidth,center}
\begin{tabular}{lccccccccccc}
\toprule
\multirow{2}{*}{\textbf{Model}} & \multirow{2}{*}{\textbf{Modality}} & \multicolumn{2}{c}{\textbf{Event Counting}} & \multicolumn{2}{c}{\textbf{Event Timing}} & \multicolumn{2}{c}{\textbf{HR Extremum Timing}} & \multicolumn{2}{c}{\textbf{Diagnosis}} & \multicolumn{2}{c}{\textbf{Evidence Reasoning}} \\
\cmidrule(lr){3-4}
\cmidrule(lr){5-6}
\cmidrule(lr){7-8}
\cmidrule(lr){9-10}
\cmidrule(lr){11-12}
& & \textbf{MAE} $\downarrow$ & \textbf{Score\textsuperscript{MAE}} $\uparrow$ & \textbf{ROUGE-L} $\uparrow$ & \textbf{F1-Bio} $\uparrow$ & \textbf{MAE ($10^{7}$ ms)} $\downarrow$ & \textbf{Score\textsuperscript{MAE}} $\uparrow$ & \textbf{ROUGE-L} $\uparrow$ & \textbf{F1-Bio} $\uparrow$ & \textbf{ROUGE-L} $\uparrow$ & \textbf{F1-Bio} $\uparrow$ \\

\hline
\rowcolor{gray!10}
\multicolumn{12}{c}{\textbf{\textit{Generalist Models}}} \\
\hline
\rowcolor{cyan!3}
GPT-5-mini & \faFont & 2.23 & 77.72 & 20.24 & 63.02 & 3.81 & 70.75 & 26.06 & 73.76 & 21.48 & 87.91 \\
\rowcolor{cyan!3}
Claude-4.5-Haiku & \faFont & 17.42 & 17.39 & 6.01 & 74.75 & 4.45 & 50.67 & 9.42 & 72.34 & 15.25 & 79.67 \\
\rowcolor{cyan!3}
Phi-4-mini-3.8B & \faFont & \underline{2.07} & \underline{78.23} & 25.20 & 81.43 & 5.43 & 21.67 & 33.07 & 83.46 & 21.19 & 87.11 \\
\rowcolor{orange!20}
\textbf{Phi-4-mini-3.8B\textsuperscript{FT}} & \faFont & \textbf{0.63} & \textbf{82.42} & \textbf{46.16} & \textbf{90.88} & \textbf{2.30} & \textbf{94.80} & \underline{81.95} & \textbf{97.91} & \underline{36.20} & \textbf{94.41} \\
\rowcolor{lime!3}
InternVL-3.5-8B & \faFilm & 16.94 & 18.70 & 17.56 & 75.61 & 4.48 & 49.67 & 25.48 & 78.89 & 18.58 & 87.91 \\
\rowcolor{lime!3}
MiniCPM-V4.5-8B & \faFilm & 14.41 & 26.86 & 11.40 & 75.88 & 4.72 & 41.71 & 12.09 & 75.16 & 18.57 & 88.32 \\
\rowcolor{lime!3}
Gemini-3.0-Flash & \faFilm & 2.48 & 76.91 & 31.99 & 81.99 & 4.46 & 50.33 & 37.74 & 79.88 & 19.47 & 85.27 \\
\rowcolor{lime!3}
Qwen3-VL-8B & \faFilm & 7.34 & 57.57 & 5.27 & 51.48 & 4.41 & 52.01 & 16.12 & 72.33 & 12.45 & 57.42 \\
\rowcolor{orange!20}
\textbf{Qwen3-VL-8B\textsuperscript{FT}} & \faFilm & \textbf{0.63} & \textbf{82.42} & \underline{44.89} & \underline{90.33} & \underline{2.74} & \underline{91.01} & \textbf{82.67} & \underline{97.67} & \textbf{38.11} & \underline{94.19} \\

\hline
\rowcolor{gray!10}
\multicolumn{12}{c}{\textbf{\textit{Medical Models}}} \\
\hline
\rowcolor{cyan!3}
LLaVA-Med-V1.5-7B & \faFont & 9.69 & 46.77 & 20.28 & 80.85 & 4.48 & 49.67 & 29.44 & 83.87 & 18.93 & 86.94 \\
\rowcolor{cyan!3}
MedGemma-1.5-4B-IT & \faFont & 7.87 & 55.16 & 18.97 & 56.99 & 4.47 & 50.00 & 18.94 & 60.97 & 14.68 & 72.65 \\
\rowcolor{cyan!3}
HealthGPT-M3-3.8B & \faFont & 9.32 & 48.48 & 11.43 & 76.49 & 4.47 & 50.00 & 12.84 & 62.94 & \underline{21.85} & 86.97 \\
\rowcolor{lime!3}
Lingshu-7B & \faFilm & 10.10 & 44.89 & 5.15 & 58.33 & 4.37 & 53.34 & 17.21 & 74.43 & 17.19 & 74.70 \\
\rowcolor{lime!3}
MedVLM-R1-2B & \faFilm & 11.39 & 39.09 & 23.56 & 80.55 & 4.49 & 49.33 & 27.10 & 72.77 & 16.73 & 79.58 \\
\rowcolor{lime!3}
HuatuoGPT-Vision-7B & \faFilm & 8.99 & 50.00 & 25.36 & 75.41 & 4.57 & 46.66 & 24.61 & 65.14 & 11.64 & 78.22 \\

\bottomrule
\end{tabular}
\end{adjustbox}
% \vspace{-3mm}
\end{table*}

Comprehensive evaluation results across \textit{Closed-QA}, \textit{Open-QA}, and \textit{Report Generation} tasks encompassing both zero-shot baselines and fine-tuned models are detailed in Tables~\ref{tab:base_closed}, \ref{tab:base_report}, and \ref{tab:base_open}, respectively.

\subsection{Baseline Evaluation}

\begin{figure}[ht]
\centering
% \vspace{-3mm}
\includegraphics[width=0.7\linewidth]{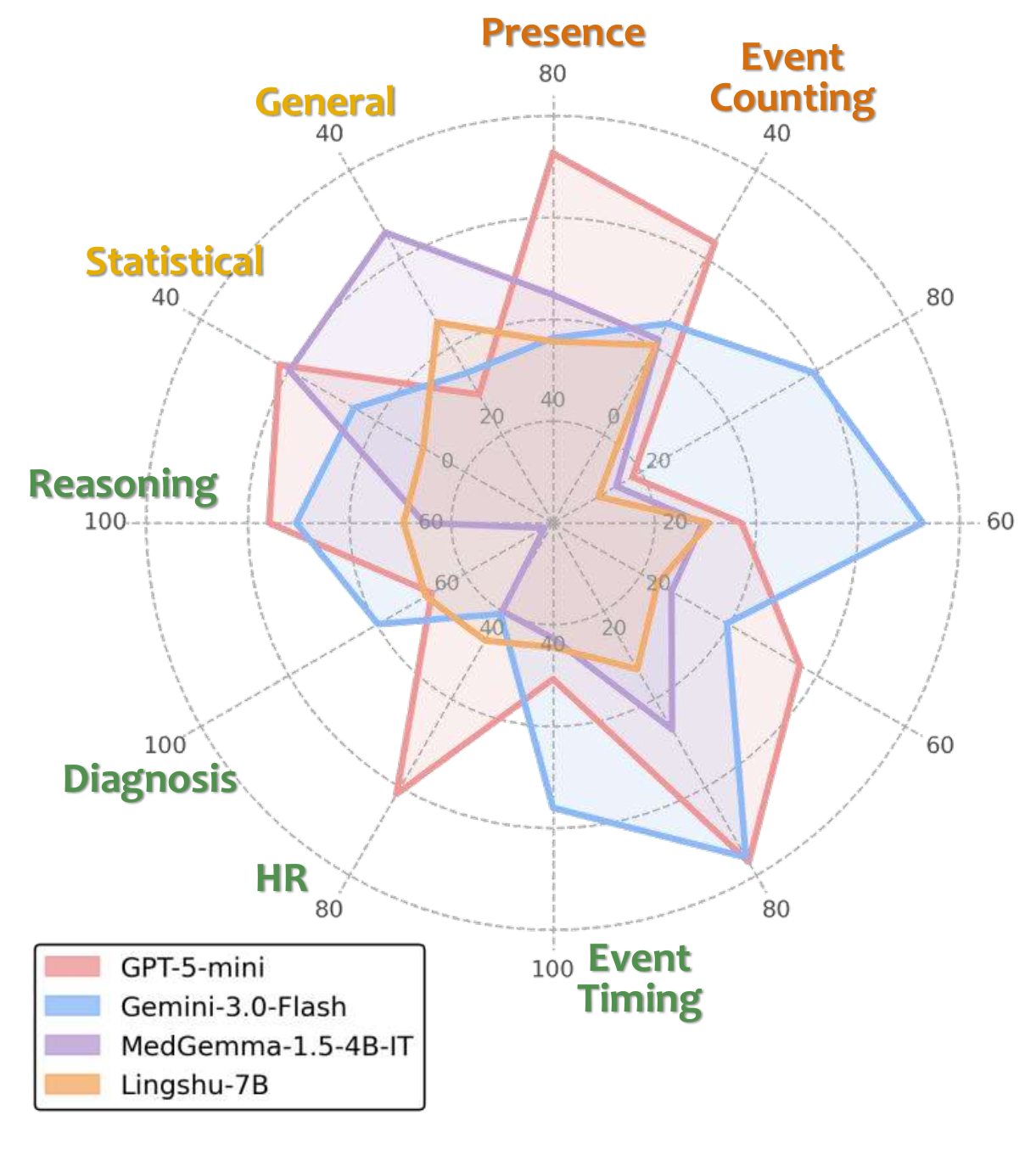}
% \vspace{-3mm}
\caption{Zero-shot performance of four representative models across diverse metrics.}
\label{fig:comparison-1}
% \vspace{-3mm}
\end{figure}

The zero-shot baseline results, as visualized in Figure~\ref{fig:comparison-1}, reveal significant limitations in existing models when processing long-term dynamic ECGs. While advanced generalist models exhibit reasonable foundational comprehension on \textit{Closed-QA} tasks---such as GPT-5-mini achieving 76.31\% accuracy in \textit{Presence} detection---performance drops sharply on fine-grained temporal problems within the \textit{Open-QA} category. Specifically, \textit{Open-QA} tasks like \textit{Event Counting} and \textit{HR Extremum Timing} prove highly challenging for most baselines. Interestingly, vision-language models like Gemini-3.0-Flash demonstrate a relative advantage on the \textit{Event Timing} sub-task of \textit{Open-QA}. Yet, all models without fine-tuning, including medical-specific ones, struggle significantly with reasoning required in \textit{Open-QA} and the ultra-long context demands of \textit{Report Generation}, yielding low entity coverage and poor \textit{ROUGE-L} scores.

\subsection{Fine-Tuning Comparison}

\begin{figure}[ht]
\centering
% \vspace{-3mm}
\includegraphics[width=0.98\linewidth]{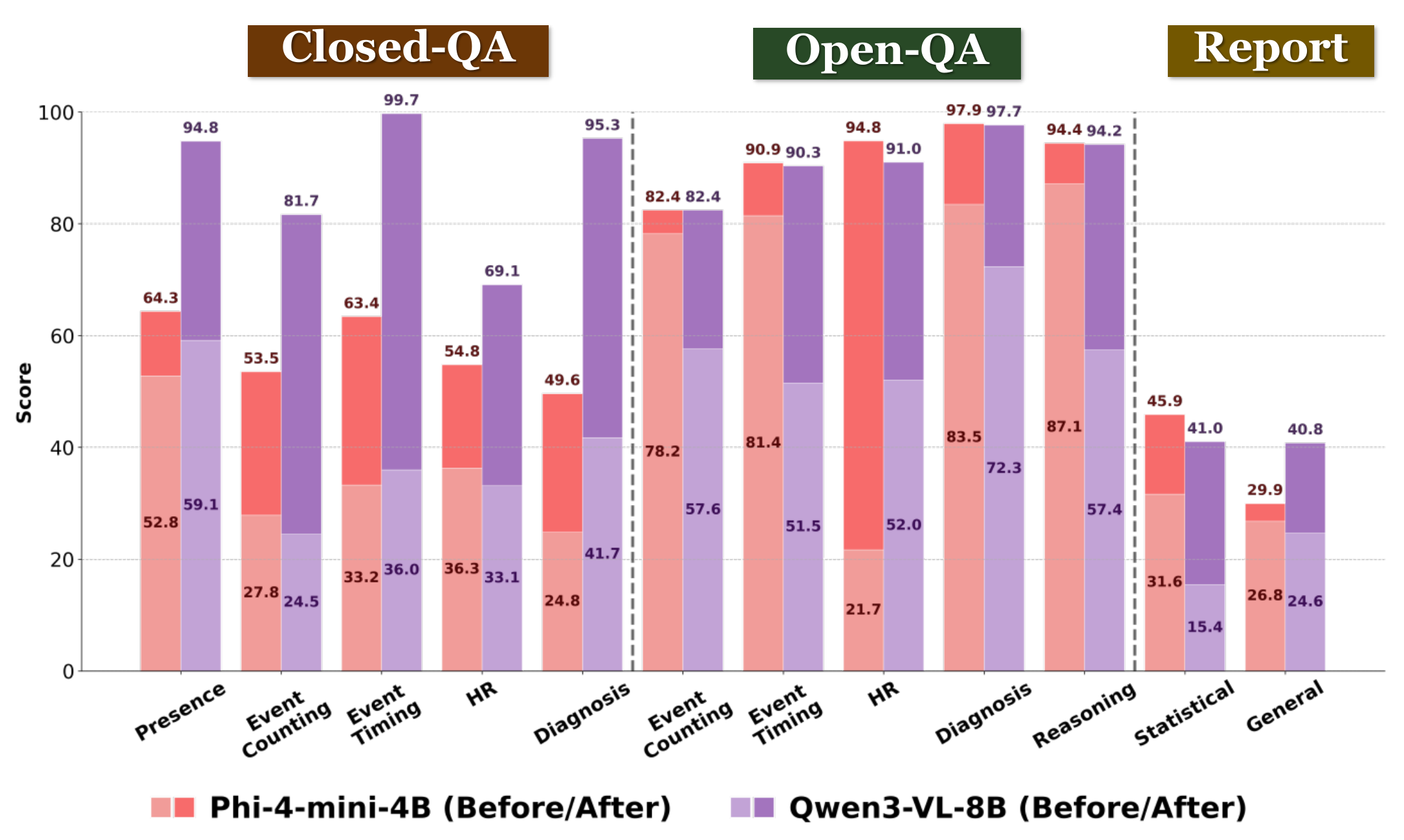}
% \vspace{-3mm}
\caption{Performance comparison of Phi-4-mini-3.8B and Qwen3-VL-8B before and after fine-tuning. Metrics include \textit{Accuracy}, \textit{Score\textsuperscript{MAE}}, and \textit{Score\textsuperscript{GPT}}. Tasks lacking these metrics are evaluated by \textit{F1-Bio}.}
\label{fig:comparison-2}
% \vspace{-3mm}
\end{figure}

Our comparative analysis of Phi-4-mini-3.8B and Qwen3-VL-8B before and after fine-tuning, shown in Figure~\ref{fig:comparison-2}, demonstrates the dataset's substantial potential. Post-tuning, both models exhibit massive performance leaps across all metrics and categories. Most notably, Qwen3-VL-8B achieved an exceptional 99.70\% accuracy on the \textit{Event Timing} sub-task of \textit{Open-QA} and 95.28\% on the \textit{Diagnosis} sub-task of \textit{Closed-QA}, while simultaneously reducing \textit{MAE} on the \textit{Event Counting} sub-task of \textit{Open-QA} from 7.34 to 0.63. Furthermore, both fine-tuned models successfully learn to synthesize ultra-long sequences, achieving state-of-the-art \textit{F1-Bio} and \textit{ROUGE-L} scores on complex \textit{Open-QA} reasoning and \textit{Report Generation} tasks. These results confirm that domain-specific multimodal alignment successfully activates the models' temporal perception and clinical reasoning capabilities, bridging the gap toward expert-level diagnostic analysis.

\section{Conclusion}

To address the challenges of MLLMs in long-term ECG analysis, we introduce \textbf{Holtercare-23K}, a large-scale dynamic ECG dataset with a tri-modal alignment, and \textbf{Holtercare-Bench}, a multimodal benchmark for evaluating long-term dynamic ECG analysis. While baselines reveal current models struggle with precise temporal localization and causal reasoning in long sequences, our fine-tuning demonstrates that high-quality multimodal data unlocks their long-context diagnostic capabilities. Looking ahead, Holtercare-Bench provides a solid foundation for long-context medical AI research. We hope our contributions accelerate the development of native dynamic ECG MLLMs.

\bibliographystyle{ACM-Reference-Format}
\balance
\bibliography{refs}

@article{bai2025qwen3,
  title={Qwen3-vl technical report},
  author={Bai, Shuai and Cai, Yuxuan and Chen, Ruizhe and Chen, Keqin and Chen, Xionghui and Cheng, Zesen and Deng, Lianghao and Ding, Wei and Gao, Chang and Ge, Chunjiang and others},
  journal={arXiv preprint arXiv:2511.21631},
  year={2025}
}

@article{wang2025internvl3,
  title={Internvl3. 5: Advancing open-source multimodal models in versatility, reasoning, and efficiency},
  author={Wang, Weiyun and Gao, Zhangwei and Gu, Lixin and Pu, Hengjun and Cui, Long and Wei, Xingguang and Liu, Zhaoyang and Jing, Linglin and Ye, Shenglong and Shao, Jie and others},
  journal={arXiv preprint arXiv:2508.18265},
  year={2025}
}

@article{yu2025minicpm,
  title={Minicpm-v 4.5: Cooking efficient mllms via architecture, data, and training recipe},
  author={Yu, Tianyu and Wang, Zefan and Wang, Chongyi and Huang, Fuwei and Ma, Wenshuo and He, Zhihui and Cai, Tianchi and Chen, Weize and Huang, Yuxiang and Zhao, Yuanqian and others},
  journal={arXiv preprint arXiv:2509.18154},
  year={2025}
}

@article{xu2025lingshu,
  title={Lingshu: A generalist foundation model for unified multimodal medical understanding and reasoning},
  author={Xu, Weiwen and Chan, Hou Pong and Li, Long and Aljunied, Mahani and Yuan, Ruifeng and Wang, Jianyu and Xiao, Chenghao and Chen, Guizhen and Liu, Chaoqun and Li, Zhaodonghui and others},
  journal={arXiv preprint arXiv:2506.07044},
  year={2025}
}

@article{li2023llava,
  title={Llava-med: Training a large language-and-vision assistant for biomedicine in one day},
  author={Li, Chunyuan and Wong, Cliff and Zhang, Sheng and Usuyama, Naoto and Liu, Haotian and Yang, Jianwei and Naumann, Tristan and Poon, Hoifung and Gao, Jianfeng},
  journal={Advances in Neural Information Processing Systems},
  volume={36},
  pages={28541--28564},
  year={2023}
}

@inproceedings{pan2025medvlm,
  title={Medvlm-r1: Incentivizing medical reasoning capability of vision-language models (vlms) via reinforcement learning},
  author={Pan, Jiazhen and Liu, Che and Wu, Junde and Liu, Fenglin and Zhu, Jiayuan and Li, Hongwei Bran and Chen, Chen and Ouyang, Cheng and Rueckert, Daniel},
  booktitle={International Conference on Medical Image Computing and Computer-Assisted Intervention},
  pages={337--347},
  year={2025},
  organization={Springer}
}

@inproceedings{lin2025healthgpt,
  title={HealthGPT: A Medical Large Vision-Language Model for Unifying Comprehension and Generation via Heterogeneous Knowledge Adaptation},
  author={Lin, Tianwei and Zhang, Wenqiao and Li, Sijing and Yuan, Yuqian and Yu, Binhe and Li, Haoyuan and He, Wanggui and Jiang, Hao and Li, Mengze and Xiaohui, Song and Tang, Siliang and Xiao, Jun and Lin, Hui and Zhuang, Yueting and Ooi, Beng Chin},
  booktitle={Proceedings of the 42nd International Conference on Machine Learning},
  pages={37975--37995},
  year={2025},
  organization={PMLR}
}

@article{sellergren2025medgemma,
  title={Medgemma technical report},
  author={Sellergren, Andrew and Kazemzadeh, Sahar and Jaroensri, Tiam and Kiraly, Atilla and Traverse, Madeleine and Kohlberger, Timo and Xu, Shawn and Jamil, Fayaz and Hughes, C{\'\i}an and Lau, Charles and others},
  journal={arXiv preprint arXiv:2507.05201},
  year={2025}
}

@inproceedings{chen2024towards,
  title={Towards injecting medical visual knowledge into multimodal llms at scale},
  author={Chen, Junying and Gui, Chi and Ouyang, Ruyi and Gao, Anningzhe and Chen, Shunian and Chen, Guiming Hardy and Wang, Xidong and Cai, Zhenyang and Ji, Ke and Wan, Xiang and others},
  booktitle={Proceedings of the 2024 conference on empirical methods in natural language processing},
  pages={7346--7370},
  year={2024}
}

@misc{claude45haiku,
  title={System Card: Claude Haiku 4.5},
  author={{Anthropic}},
  howpublished={\url{https://www.anthropic.com/system-cards}},
  year={2025}
}

@misc{gpt5,
  title={GPT-5 System Card},
  author={{OpenAI}},
  howpublished={\url{https://cdn.openai.com/gpt-5-system-card.pdf}},
  year={2025}
}

@misc{gemini3flash,
  title={Gemini 3 Flash Model Card},
  author={{Google DeepMind}},
  howpublished={\url{https://deepmind.google/models/model-cards/gemini-3-flash/}},
  year={2025}
}

@article{abdin2024phi,
  title={Phi-4 technical report},
  author={Abdin, Marah and Aneja, Jyoti and Behl, Harkirat and Bubeck, S{\'e}bastien and Eldan, Ronen and Gunasekar, Suriya and Harrison, Michael and Hewett, Russell J and Javaheripi, Mojan and Kauffmann, Piero and others},
  journal={arXiv preprint arXiv:2412.08905},
  year={2024}
}

@article{PhysioNet-mimic-iv-ecg-1.0,
  author = {Gow, Brian and Pollard, Tom and Nathanson, Larry A and Johnson, Alistair and Moody, Benjamin and Fernandes, Chrystinne and Greenbaum, Nathaniel and Waks, Jonathan W and Eslami, Parastou and Carbonati, Tanner and Chaudhari, Ashish and Herbst, Elizabeth and Moukheiber, Dana and Berkowitz, Seth and Mark, Roger and Horng, Steven},
  title = {{MIMIC-IV-ECG: Diagnostic Electrocardiogram Matched Subset}},
  journal = {{PhysioNet}},
  year = {2023},
  month = sep,
  note = {Version 1.0},
  doi = {10.13026/4nqg-sb35},
  url = {https://doi.org/10.13026/4nqg-sb35}
}

@article{PhysioNet-ptb-xl-1.0.3,
  author = {Wagner, Patrick and Strodthoff, Nils and Bousseljot, Ralf-Dieter and Samek, Wojciech and Schaeffter, Tobias},
  title = {{PTB-XL, a large publicly available electrocardiography dataset}},
  journal = {{PhysioNet}},
  year = {2022},
  month = nov,
  note = {Version 1.0.3},
  doi = {10.13026/kfzx-aw45},
  url = {https://doi.org/10.13026/kfzx-aw45}
}

@article{PhysioNet-ecg-arrhythmia-1.0.0,
  author = {Zheng, Jianwei and Guo, Hangyuan and Chu, Huimin},
  title = {{A large scale 12-lead electrocardiogram database for arrhythmia study}},
  journal = {{PhysioNet}},
  year = {2022},
  month = aug,
  note = {Version 1.0.0},
  doi = {10.13026/wgex-er52},
  url = {https://doi.org/10.13026/wgex-er52}
}

@article{PhysioNet-ludb-1.0.1,
  author = {Kalyakulina, Alena and Yusipov, Igor and Moskalenko, Viktor and Nikolskiy, Alexander and Kosonogov, Konstantin and Zolotykh, Nikolai and Ivanchenko, Mikhail},
  title = {{Lobachevsky University Electrocardiography Database}},
  journal = {{PhysioNet}},
  year = {2021},
  month = jan,
  note = {Version 1.0.1},
  doi = {10.13026/eegm-h675},
  url = {https://doi.org/10.13026/eegm-h675}
}

@article{PhysioNet-icentia11k-continuous-ecg-1.0,
  author = {Tan, Shawn and Ortiz-Gagné, Satya and Beaudoin-Gagnon, Nicolas and Fecteau, Pierre and Courville, Aaron and Bengio, Yoshua and Cohen, Joseph Paul},
  title = {{Icentia11k Single Lead Continuous Raw Electrocardiogram Dataset}},
  journal = {{PhysioNet}},
  year = {2022},
  month = apr,
  note = {Version 1.0},
  doi = {10.13026/kk0v-r952},
  url = {https://doi.org/10.13026/kk0v-r952}
}

@article{incartdb,
  author = {Yakushenko, Evgeny},
  title = {{St Petersburg INCART 12-lead Arrhythmia Database}},
  journal = {{PhysioNet}},
  year = {2008},
  month = may,
  note = {Version 1.0.0},
  doi = {10.13026/C2V88N},
  url = {https://doi.org/10.13026/C2V88N}
}

@article{ltdb,
  author = {Moody, George B and Mark, Roger G},
  title = {{MIT-BIH Long-Term ECG Database}},
  journal = {{PhysioNet}},
  year = {1999},
  month = aug,
  note = {Version 1.0.0},
  doi = {10.13026/C2KS3F},
  url = {https://doi.org/10.13026/C2KS3F}
}

@article{nsrdb,
  author = {Moody, George B and Mark, Roger G},
  title = {{MIT-BIH Normal Sinus Rhythm Database}},
  journal = {{PhysioNet}},
  year = {1999},
  month = aug,
  note = {Version 1.0.0},
  doi = {10.13026/C2NK5R},
  url = {https://doi.org/10.13026/C2NK5R}
}

@article{PhysioNet-shdb-af-1.0.1,
  author = {Tsutsui, Kenta and {Biton Brimer}, Shany and Behar, Joachim},
  title = {{SHDB-AF: a Japanese Holter ECG database of atrial fibrillation}},
  journal = {{PhysioNet}},
  year = {2025},
  month = apr,
  note = {Version 1.0.1},
  doi = {10.13026/n6yq-fq90},
  url = {https://doi.org/10.13026/n6yq-fq90}
}

@article{moody2001impact,
  title={The impact of the MIT-BIH arrhythmia database},
  author={Moody, George B and Mark, Roger G},
  journal={IEEE engineering in medicine and biology magazine},
  volume={20},
  number={3},
  pages={45--50},
  year={2001},
  publisher={IEEE}
}

@article{jager2003long,
  title={Long-term ST database: a reference for the development and evaluation of automated ischaemia detectors and for the study of the dynamics of myocardial ischaemia},
  author={Jager, Franc and Taddei, Alessandro and Moody, George B and Emdin, Michele and Antoli{\v{c}}, G and Dorn, Roman and Smrdel, Ales and Marchesi, Carlo and Mark, Roger G},
  journal={Medical and Biological Engineering and Computing},
  volume={41},
  number={2},
  pages={172--182},
  year={2003},
  publisher={Springer}
}

@article{moody1983new,
  title={A new method for detecting atrial fibrillation using RR intervals},
  author={Moody, George B and Mark, Roger G},
  journal={Proc. Comput. Cardiol.},
  volume={10},
  pages={227--230},
  year={1983}
}

@phdthesis{greenwald1986development,
  title={The development and analysis of a ventricular fibrillation detector},
  author={Greenwald, Scott David},
  year={1986},
  school={Massachusetts Institute of Technology}
}

@article{petrutiu2007abrupt,
  title={Abrupt changes in fibrillatory wave characteristics at the termination of paroxysmal atrial fibrillation in humans},
  author={Petrutiu, Simona and Sahakian, Alan V and Swiryn, Steven},
  journal={Europace},
  volume={9},
  number={7},
  pages={466--470},
  year={2007},
  publisher={Oxford University Press}
}

@inproceedings{laguna1997database,
  title={A database for evaluation of algorithms for measurement of QT and other waveform intervals in the ECG},
  author={Laguna, Pablo and Mark, Roger G and Goldberg, A and Moody, George B},
  booktitle={Computers in cardiology 1997},
  pages={673--676},
  year={1997},
  organization={IEEE}
}

@inproceedings{penzel2000apnea,
  title={The apnea-ECG database},
  author={Penzel, Thomas and Moody, George B and Mark, Roger G and Goldberger, Ary L and Peter, J Hermann},
  booktitle={Computers in Cardiology 2000. Vol. 27 (Cat. 00CH37163)},
  pages={255--258},
  year={2000},
  organization={IEEE}
}

@article{liu2018open,
  title={An open access database for evaluating the algorithms of electrocardiogram rhythm and morphology abnormality detection},
  author={Liu, Feifei and Liu, Chengyu and Zhao, Lina and Zhang, Xiangyu and Wu, Xiaoling and Xu, Xiaoyan and Liu, Yulin and Ma, Caiyun and Wei, Shoushui and He, Zhiqiang and others},
  journal={Journal of Medical Imaging and Health Informatics},
  volume={8},
  number={7},
  pages={1368--1373},
  year={2018},
  publisher={American Scientific Publishers}
}

@article{makowski2021neurokit2,
  title={NeuroKit2: A Python toolbox for neurophysiological signal processing},
  author={Makowski, Dominique and Pham, Tam and Lau, Zen J and Brammer, Jan C and Lespinasse, Fran{\c{c}}ois and Pham, Hung and Sch{\"o}lzel, Christopher and Chen, SH Annabel},
  journal={Behavior research methods},
  volume={53},
  number={4},
  pages={1689--1696},
  year={2021},
  publisher={Springer}
}

@article{hunter2007matplotlib,
  title={Matplotlib: A 2D graphics environment},
  author={Hunter, John D},
  journal={Computing in science \& engineering},
  volume={9},
  number={3},
  pages={90--95},
  year={2007},
  publisher={IEEE}
}

@inproceedings{lin2004rouge,
  title={Rouge: A package for automatic evaluation of summaries},
  author={Lin, Chin-Yew},
  booktitle={Text summarization branches out},
  pages={74--81},
  year={2004}
}

@inproceedings{papineni2002bleu,
  title={Bleu: a method for automatic evaluation of machine translation},
  author={Papineni, Kishore and Roukos, Salim and Ward, Todd and Zhu, Wei-Jing},
  booktitle={Proceedings of the 40th annual meeting of the Association for Computational Linguistics},
  pages={311--318},
  year={2002}
}

@inproceedings{ramshaw1995text,
  title={Text chunking using transformation-based learning},
  author={Ramshaw, Lance and Marcus, Mitch},
  booktitle={Third workshop on very large corpora},
  year={1995}
}

@article{gramfort2014mne,
  title={MNE software for processing MEG and EEG data},
  author={Gramfort, Alexandre and Luessi, Martin and Larson, Eric and Engemann, Denis A and Strohmeier, Daniel and Brodbeck, Christian and Parkkonen, Lauri and H{\"a}m{\"a}l{\"a}inen, Matti S},
  journal={neuroimage},
  volume={86},
  pages={446--460},
  year={2014},
  publisher={Elsevier}
}

@inproceedings{moor2023med,
  title={Med-flamingo: a multimodal medical few-shot learner},
  author={Moor, Michael and Huang, Qian and Wu, Shirley and Yasunaga, Michihiro and Dalmia, Yash and Leskovec, Jure and Zakka, Cyril and Reis, Eduardo Pontes and Rajpurkar, Pranav},
  booktitle={Machine learning for health (ML4H)},
  pages={353--367},
  year={2023},
  organization={PMLR}
}

@article{li2025electrocardiogram,
  title={An electrocardiogram foundation model built on over 10 million recordings},
  author={Li, Jun and Aguirre, Aaron D and Junior, Valdery Moura and Jin, Jiarui and Liu, Che and Zhong, Lanhai and Sun, Chenxi and Clifford, Gari and Brandon Westover, M and Hong, Shenda},
  journal={Nejm ai},
  volume={2},
  number={7},
  pages={AIoa2401033},
  year={2025},
  publisher={Massachusetts Medical Society}
}

@article{mckeen2025ecg,
  title={Ecg-fm: An open electrocardiogram foundation model},
  author={McKeen, Kaden and Masood, Sameer and Toma, Augustin and Rubin, Barry and Wang, Bo},
  journal={Jamia Open},
  volume={8},
  number={5},
  pages={ooaf122},
  year={2025},
  publisher={Oxford University Press}
}

@article{tian2024foundation,
  title={Foundation model of ECG diagnosis: Diagnostics and explanations of any form and rhythm on ECG},
  author={Tian, Yuanyuan and Li, Zhiyuan and Jin, Yanrui and Wang, Mengxiao and Wei, Xiaoyang and Zhao, Liqun and Liu, Yunqing and Liu, Jinlei and Liu, Chengliang},
  journal={Cell Reports Medicine},
  volume={5},
  number={12},
  year={2024},
  publisher={Elsevier}
}

@inproceedings{wan2025meit,
  title={MEIT: Multimodal electrocardiogram instruction tuning on large language models for report generation},
  author={Wan, Zhongwei and Liu, Che and Wang, Xin and Tao, Chaofan and Shen, Hui and Xiong, Jing and Arcucci, Rossella and Yao, Huaxiu and Zhang, Mi},
  booktitle={Findings of the association for computational linguistics: ACL 2025},
  pages={14510--14527},
  year={2025}
}

@article{liu2024teach,
  title={Teach multimodal llms to comprehend electrocardiographic images},
  author={Liu, Ruoqi and Bai, Yuelin and Yue, Xiang and Zhang, Ping},
  journal={arXiv preprint arXiv:2410.19008},
  year={2024}
}

@article{han2024ecg,
  title={Ecg-byte: A tokenizer for end-to-end generative electrocardiogram language modeling},
  author={Han, William and Duan, Chaojing and Rosenberg, Michael A and Liu, Emerson and Zhao, Ding},
  journal={arXiv preprint arXiv:2412.14373},
  year={2024}
}

@article{liu2024zero,
  title={Zero-shot ecg classification with multimodal learning and test-time clinical knowledge enhancement},
  author={Liu, Che and Wan, Zhongwei and Ouyang, Cheng and Shah, Anand and Bai, Wenjia and Arcucci, Rossella},
  journal={arXiv preprint arXiv:2403.06659},
  year={2024}
}

@article{lau2018dataset,
  title={A dataset of clinically generated visual questions and answers about radiology images},
  author={Lau, Jason J and Gayen, Soumya and Ben Abacha, Asma and Demner-Fushman, Dina},
  journal={Scientific data},
  volume={5},
  number={1},
  pages={180251},
  year={2018},
  publisher={Nature Publishing Group}
}

@inproceedings{liu2021slake,
  title={Slake: A semantically-labeled knowledge-enhanced dataset for medical visual question answering},
  author={Liu, Bo and Zhan, Li-Ming and Xu, Li and Ma, Lin and Yang, Yan and Wu, Xiao-Ming},
  booktitle={2021 IEEE 18th international symposium on biomedical imaging (ISBI)},
  pages={1650--1654},
  year={2021},
  organization={IEEE}
}

@inproceedings{fu2025video,
  title={Video-mme: The first-ever comprehensive evaluation benchmark of multi-modal llms in video analysis},
  author={Fu, Chaoyou and Dai, Yuhan and Luo, Yongdong and Li, Lei and Ren, Shuhuai and Zhang, Renrui and Wang, Zihan and Zhou, Chenyu and Shen, Yunhang and Zhang, Mengdan and others},
  booktitle={Proceedings of the IEEE/CVF conference on computer vision and pattern recognition},
  pages={24108--24118},
  year={2025}
}

@inproceedings{wang2025lvbench,
  title={Lvbench: An extreme long video understanding benchmark},
  author={Wang, Weihan and He, Zehai and Hong, Wenyi and Cheng, Yean and Zhang, Xiaohan and Qi, Ji and Ding, Ming and Gu, Xiaotao and Huang, Shiyu and Xu, Bin and others},
  booktitle={Proceedings of the IEEE/CVF International Conference on Computer Vision},
  pages={22958--22967},
  year={2025}
}

@article{chaves2024towards,
  title={Towards a clinically accessible radiology foundation model: open-access and lightweight, with automated evaluation},
  author={Chaves, Juan Manuel Zambrano and Huang, Shih-Cheng and Xu, Yanbo and Xu, Hanwen and Usuyama, Naoto and Zhang, Sheng and Wang, Fei and Xie, Yujia and Khademi, Mahmoud and Yang, Ziyi and others},
  journal={arXiv preprint arXiv:2403.08002},
  year={2024}
}

@article{zhou2023skingpt,
  title={SkinGPT-4: an interactive dermatology diagnostic system with visual large language model},
  author={Zhou, Juexiao and He, Xiaonan and Sun, Liyuan and Xu, Jiannan and Chen, Xiuying and Chu, Yuetan and Zhou, Longxi and Liao, Xingyu and Zhang, Bin and Gao, Xin},
  journal={arXiv preprint arXiv:2304.10691},
  year={2023}
}

@article{yu2023ecg,
  title={ECG-SL: electrocardiogram (ECG) segment learning, a deep learning method for ECG signal},
  author={Yu, Han and Yang, Huiyuan and Sano, Akane},
  journal={arXiv preprint arXiv:2310.00818},
  year={2023}
}

@article{jin2025reading,
  title={Reading your heart: Learning ecg words and sentences via pre-training ecg language model},
  author={Jin, Jiarui and Wang, Haoyu and Li, Hongyan and Li, Jun and Pan, Jiahui and Hong, Shenda},
  journal={arXiv preprint arXiv:2502.10707},
  year={2025}
}

@article{yang2025ecg,
  title={ECG-LM: understanding electrocardiogram with a large language model},
  author={Yang, Kai and Hong, Massimo and Zhang, Jiahuan and Luo, Yizhen and Zhao, Suyuan and Zhang, Ou and Yu, Xiaomao and Zhou, Jiawen and Yang, Liuqing and Zhang, Ping and others},
  journal={Health Data Science},
  volume={5},
  pages={0221},
  year={2025},
  publisher={AAAS}
}

@article{cai2025supreme,
  title={SuPreME: A Supervised Pre-training Framework for Multimodal ECG Representation Learning},
  author={Cai, Mingsheng and Jiang, Jiuming and Huang, Wenhao and Liu, Che and Arcucci, Rossella},
  journal={arXiv preprint arXiv:2502.19668},
  volume={3},
  year={2025}
}

@article{yu2024ecg,
  title={Ecg semantic integrator (esi): A foundation ecg model pretrained with llm-enhanced cardiological text},
  author={Yu, Han and Guo, Peikun and Sano, Akane},
  journal={arXiv preprint arXiv:2405.19366},
  year={2024}
}

@article{zhang2023pmc,
  title={Pmc-vqa: Visual instruction tuning for medical visual question answering},
  author={Zhang, Xiaoman and Wu, Chaoyi and Zhao, Ziheng and Lin, Weixiong and Zhang, Ya and Wang, Yanfeng and Xie, Weidi},
  journal={arXiv preprint arXiv:2305.10415},
  year={2023}
}

@article{chen2024gmai,
  title={Gmai-mmbench: A comprehensive multimodal evaluation benchmark towards general medical ai},
  author={Chen, Pengcheng and Ye, Jin and Wang, Guoan and Li, Yanjun and Deng, Zhongying and Li, Wei and Li, Tianbin and Duan, Haodong and Huang, Ziyan and Su, Yanzhou and others},
  journal={Advances in Neural Information Processing Systems},
  volume={37},
  pages={94327--94427},
  year={2024}
}

@article{oh2023ecg,
  title={Ecg-qa: A comprehensive question answering dataset combined with electrocardiogram},
  author={Oh, Jungwoo and Lee, Gyubok and Bae, Seongsu and Kwon, Joon-myoung and Choi, Edward},
  journal={Advances in Neural Information Processing Systems},
  volume={36},
  pages={66277--66288},
  year={2023}
}

@inproceedings{li2025eyecaregpt,
  title={EyecareGPT: Boosting Comprehensive Ophthalmology Understanding with Tailored Dataset, Benchmark and Model},
  author={Li, Sijing and Lin, Tianwei and Lin, Lingshuai and Zhang, Wenqiao and Liu, Jiang and Yang, Xiaoda and Li, Juncheng and He, Yucheng and Song, Xiaohui and Xiao, Jun and Zhuang, Yueting and Ooi, Beng Chin},
  booktitle={Proceedings of the 33rd ACM International Conference on Multimedia},
  pages={3893--3902},
  year={2025}
}

@inproceedings{lin2026omnict,
  title={OmniCT: Towards a Unified Slice-Volume LVLM for Comprehensive CT Analysis},
  author={Lin, Tianwei and Qiu, Zhongwei and Zhang, Wenqiao and Liu, Jiang and Xie, Yihan and Gao, Mingjian and Fan, Zhenxuan and Li, Zhaocheng and Li, Sijing and Xie, Zhongle and Lu, Peng and Zhuang, Yueting and Zhang, Ling and Ooi, Beng Chin and Xia, Yingda},
  booktitle={International Conference on Learning Representations},
  year={2026}
}

@inproceedings{li2026tumorchain,
  title={TumorChain: Interleaved Multimodal Chain-of-Thought Reasoning for Traceable Clinical Tumor Analysis},
  author={Li, Sijing and Qiu, Zhongwei and Liu, Jiang and Zhang, Wenqiao and Lin, Tianwei and Xie, Yihan and An, Jianxiang and Yun, Boxiang and Yang, Chenglin and Xiao, Jun and Guo, Guangyu and Yao, Jiawen and Liu, Wei and Gao, Yuan and Yan, Ke and Cao, Weiwei and Zheng, Zhilin and Mok, Tony C. W. and Cao, Kai and Shi, Yu and Zhang, Jiuyu and Zhou, Jian and Ooi, Beng Chin and Xia, Yingda and Zhang, Ling},
  booktitle={International Conference on Learning Representations},
  year={2026}
}

@inproceedings{lin2026regulating,
  title={Regulating Anatomy-Aware Rewards via Trajectory-Integral Feedback for Volumetric Computed Tomography Analysis},
  author={Lin, Tianwei and Qiu, Zhongwei and Cao, Jie and Liu, Jiang and Yan, Wenjie and Zhang, Bo and Zhong, Yu and Zhang, Wenqiao and Xia, Yingda and Zhang, Ling},
  booktitle={Proceedings of the 43rd International Conference on Machine Learning},
  volume={306},
  series={Proceedings of Machine Learning Research},
  year={2026},
  publisher={PMLR}
}

@inproceedings{li2026emrl,
  title={E-MRL: Cross-view Aligned Evidence-driven Multimodal Reinforcement Learning for Reliable 3D Tumor Analysis},
  author={Li, Sijing and Qiu, Zhongwei and Wang, Zhuoya and Yun, Boxiang and Yi, Zhenyu and Xu, Jianwei and Zhang, Wenqiao and Xia, Yingda and Zhang, Ling},
  booktitle={MICCAI 2026},
  year={2026}
}

@inproceedings{yuan2025videorefer,
  title={Videorefer suite: Advancing spatial-temporal object understanding with video llm},
  author={Yuan, Yuqian and Zhang, Hang and Li, Wentong and Cheng, Zesen and Zhang, Boqiang and Li, Long and Li, Xin and Zhao, Deli and Zhang, Wenqiao and Zhuang, Yueting and others},
  booktitle={2025 IEEE/CVF Conference on Computer Vision and Pattern Recognition (CVPR)},
  pages={18970--18980},
  year={2025},
  organization={IEEE}
}

@article{yuan2025pixelrefer,
  title={Pixelrefer: A unified framework for spatio-temporal object referring with arbitrary granularity},
  author={Yuan, Yuqian and Zhang, Wenqiao and Li, Xin and Wang, Shihao and Li, Kehan and Li, Wentong and Xiao, Jun and Zhang, Lei and Ooi, Beng Chin},
  journal={arXiv preprint arXiv:2510.23603},
  year={2025}
}

@article{yuan2026instructsam,
  title={InstructSAM: Segment Any Instance with Any Instructions},
  author={Yuan, Yuqian and Li, Wentong and Li, Zhaocheng and Lin, Yutong and Li, Juncheng and Tang, Siliang and Xiao, Jun and Zhuang, Yueting and Zhang, Wenqiao},
  journal={arXiv preprint arXiv:2605.26102},
  year={2026}
}

@article{zhang2024hyperllava,
  title={Hyperllava: Dynamic visual and language expert tuning for multimodal large language models},
  author={Zhang, Wenqiao and Lin, Tianwei and Liu, Jiang and Shu, Fangxun and Li, Haoyuan and Zhang, Lei and Wanggui, He and Zhou, Hao and Lv, Zheqi and Jiang, Hao and others},
  journal={arXiv preprint arXiv:2403.13447},
  year={2024}
}

@article{gao2026visualthink,
  title={VisualThink-VLA: Visual Intermediate Reasoning for Effective and Low-Latency Vision-Language-Action Policies},
  author={Gao, Mingjian and Zhang, Wenqiao and Yuan, Yuqian and Dai, Yang and Yu, Binhe and Lv, Zheqi and Zheng, Haoyu and Zhu, Jiaqi and Ge, Zhiqi and Wan, Zixuan and others},
  journal={arXiv preprint arXiv:2605.30011},
  year={2026}
}

\appendix

\cleardoublepage
\onecolumn

\begin{center}
% \vspace{5mm}
\huge \textbf{Appendix}
\vspace{3mm}
\end{center}

This is the appendix for ``Holtercare-Bench: A Multimodal Benchmark for Evaluating Long-Term Dynamic ECG Analysis''.

This appendix is organized as follows:

\begin{itemize}
\item Section~\ref{app:ann} provides the detailed distribution and frequencies of the clinical annotations within Holtercare-23K.
\item Section~\ref{app:prompt} presents the specific prompt templates utilized for LLM-based QA generation and report evaluation.
\item Section~\ref{app:result} details supplemental generation metrics, \textit{BLEU} n-gram overlap metrics, and additional analyses on modality alignment and statistical significance of fine-tuning improvements.
\end{itemize}

\section{Detailed Distribution of Clinical Annotations}
\label{app:ann}

While Figure~\ref{fig:count} contains an overview of the most frequent categories, to provide a deeper understanding of the scale and clinical diversity of Holtercare-23K, we present the exhaustive frequencies of our expert-verified beat and rhythm annotations.

Table~\ref{tab:beat} details the occurrence counts for all beat-level annotations. This includes a massive scale of valid QRS complexes---such as normal beats (\textit{N}), atrial fibrillation beats (\textit{Af}), and ventricular premature beats (\textit{V})---along with non-QRS elements and artifacts.

\begin{table*}[ht]
\centering
% \small
% \renewcommand{\arraystretch}{1.2}
\caption{Frequency of beat-level annotations across Holtercare-23K.}
\label{tab:beat}
% \begin{adjustbox}{width=1.2\linewidth,center}
\begin{tabular}{lll}
\toprule
\textbf{Label} & \textbf{Description} & \textbf{Count} \\

\hline
\rowcolor{gray!10} 
\multicolumn{3}{c}{\textbf{\textit{Valid QRS Complexes}}} \\
\hline
N & Normal Beat & 71,429,269 \\
Af & Atrial Fibrillation & 4,822,970 \\
S & Atrial Premature Beat (APB) & 1,042,095 \\
P & Unclassified Pacing & 840,298 \\
V & Ventricular Premature Beat (VPB) & 779,227 \\
Se & Atrial Escape Beat & 587,037 \\
AF & Atrial Flutter & 490,579 \\
Je & Junctional Escape Beat & 179,605 \\
B & Bundle Branch Block & 2,051 \\
Ve & Ventricular Escape Beat & 1,969 \\
J & Junctional Premature Beat & 182 \\
? & Questionable Beat & 30 \\
F & Fusion Beat & 1 \\
aP & Atrial Single-Chamber Pacing & 0 \\
vP & Ventricular Single-Chamber Pacing & 0 \\
dP & Dual-Chamber Pacing & 0 \\
Ab & APB with Aberrant Ventricular Conduction & 0 \\
Va & Aberrant Ventricular Conduction & 0 \\

\hline
\rowcolor{gray!10} 
\multicolumn{3}{c}{\textbf{\textit{Non-QRS Elements}}} \\
\hline
X & Artifact & 833,824 \\
Sa & Non-conducted APB & 32,645 \\
p & P-wave & 0 \\
t & T-wave & 0 \\

\bottomrule
\end{tabular}
% \end{adjustbox}
\end{table*}

Table~\ref{tab:rhythm} outlines the comprehensive frequencies of the rhythm-level annotation. These capture complex, continuous cardiac events and diverse arrhythmias ranging from common premature atrial contractions (PAC) to severe, life-threatening events like ventricular fibrillation (VF) and asystole.

\begin{table*}[ht]
\centering
% \small
% \renewcommand{\arraystretch}{1.2}
\caption{Frequency of rhythm-level annotations across Holtercare-23K.}
\label{tab:rhythm}
% \begin{adjustbox}{width=1.8\linewidth,center}
\begin{tabular}{ll|ll}
\toprule
\textbf{Event} & \textbf{Count} & \textbf{Event} & \textbf{Count} \\
\midrule
Premature Atrial Contraction (PAC) & 888 & Sinus Arrest & 6 \\
Premature Ventricular Contraction (PVC) & 775 & Atrial Flutter (AFL) & 6 \\
PAC Couplets & 396 & Atrial Escape Rhythm & 4 \\
Atrial Tachycardia (AT) & 339 & Wandering Atrial Pacemaker (WAP) & 4 \\
PVC Couplets & 152 & Ventricular Escape Rhythm & 3 \\
Ventricular Tachycardia (VT) & 94 & Atrial Undersensing & 2 \\
PAC Bigeminy & 82 & T-Wave Alternans (TWA) & 2 \\
PAC Trigeminy & 64 & Ventricular Capture Management (VCM) & 3 \\
PVC Trigeminy & 58 & Premature Junctional Contraction (PJC) & 2 \\
PVC Bigeminy & 55 & Atrial Capture Management (ACM) & 2 \\
Asystole & 44 & Defibrillation & 2 \\
Sinus Arrhythmia & 44 & ST Segment Changes & 2 \\
Long R-R Interval & 33 & Accelerated Atrial Escape Rhythm & 1 \\
Ventricular Fibrillation (VF) & 23 & Anti-Tachycardia Pacing (ATP) & 1 \\
Second-Degree Atrioventricular (AV) Block & 15 & Ventricular Dissociation & 1 \\
Junctional Escape Rhythm & 12 & OptiVol Fluid Status & 1 \\
First-Degree Atrioventricular (AV) Block & 9 & Supraventricular Tachycardia (SVT) & 1 \\
Accelerated Idioventricular Rhythm (AIVR) & 6 & Ventricular Sense Response (VSR) & 1 \\
Ventricular Flutter (VFL) & 6 & Chest Compression Waveform & 1 \\
Atrial Fibrillation (AF) & 6 \\
\bottomrule
\end{tabular}
% \end{adjustbox}
\end{table*}

\section{Prompt for QA Generation and Report Evaluation}
\label{app:prompt}

As outlined in our methodology, we use GPT-5-mini~\cite{gpt5} for both the automated construction of complex reasoning tasks and the LLM-as-a-judge evaluation of long-context reports. To ensure transparency and reproducibility, we provide the exact representative prompts used in these two pipelines.

\noindent \textbf{QA Generation.} Figure~\ref{fig:prompt_gen} displays a representative structured prompt utilized by HolterAgent to generate \textit{Open-QA} pairs, specifically showcasing the \textit{Evidence Reasoning} sub-task. By feeding the model explicit patient information, the prompt strictly forces the LLM to extract timing, rhythm, and morphology clues directly from the provided text, effectively mitigating hallucinated waveform features.

\definecolor{titlebar}{RGB}{200,200,200}
\definecolor{boxbg}{RGB}{255,255,255}

\begin{figure*}[ht]
\centering
\begin{adjustbox}{width=0.95\textwidth}
\begin{tcolorbox}[
enhanced,
colback=boxbg,
colframe=black,
arc=0pt,
outer arc=0pt,
boxrule=1pt,
toprule=1.5pt,
bottomrule=1pt,
leftrule=1pt,
rightrule=1pt,
titlerule=0pt,
title={\color{black}\large\textbf{QA Generation Prompt}},
fonttitle=\bfseries,
attach boxed title to top left={xshift=5mm, yshift=-3mm},
boxed title style={
colback=titlebar,
colframe=titlebar,
arc=0pt,
outer arc=0pt,
boxrule=0pt,
toprule=0pt,
bottomrule=0pt,
leftrule=0pt,
rightrule=0pt,
}
]

\vspace{5mm}

Your task is to generate high-fidelity \textit{Open-QA} data in \textit{Evidence Reasoning} tasks for Holter ECG diagnostic training. Given a patient's structured information, you must generate exactly one QA pair and construct a rigorous, evidence-based clinical reasoning process that verifies the target diagnosis based solely on the provided inputs. Zero hallucination is tolerated. Adhere strictly to requirements below.

\vspace{5mm}

\textbf{Patient Information}
\begin{itemize}
\item \textbf{Age:} \texttt{\{\{AGE\}\}}
\item \textbf{Gender:} \texttt{\{\{GENDER\}\}}
\item \textbf{Electronic Medical Records:} \texttt{\{\{EMR\}\}}
\item \textbf{Beat-Level Annotations:} \texttt{\{\{BEAT\_ANNOTATIONS\}\}}
\item \textbf{Rhythm-Level Annotations:} \texttt{\{\{RHYTHM\_ANNOTATIONS\}\}}
\item \textbf{Report-Level Summaries:} \texttt{\{\{REPORT\_SUMMARY\}\}}
\item \textbf{Target Disease}: \texttt{\{\{TARGET\_DISEASE\}\}}
\end{itemize}

\vspace{5mm}

\textbf{Requirements}
\begin{enumerate}
\item Requirements for \textbf{questions}:
    \begin{itemize}
        \item Introduce the patient's \textbf{age} and \textbf{gender} (e.g., \textit{``The current monitoring segment comes from a 61-year-old male.''}), explicitly mention the exact \textbf{target disease}, and ask whether it is present on the visible ECG waveform.
        \item Ask the reader to specify which waveform features support that determination (e.g., prompting for rhythm clues, timing, P-wave morphology, PR-interval behavior, or QRS characteristics).
        \item Vary the phrasing and sentence structure naturally, but \textbf{do not change the core verification task.}
    \end{itemize}

\item Requirements for \textbf{answers}:
    \begin{itemize}
        \item Begin by directly confirming the presence of the target disease (e.g., \textit{``Yes, \{\{TARGET\_DISEASE\}\} is present.''} or \textit{``Confirmed.''}).
        \item Provide a structured clinical reasoning process by summarizing clues extracted from the provided annotations and summaries. Organize the reasoning into logical clinical dimensions where applicable, such as:
        \begin{itemize}
            \item \textbf{Timing / Rhythm Clues:} Discuss rates, RR intervals, coupling intervals, and event burden based on the report and beat/rhythm Annotations.
            \item \textbf{Morphology Clues:} Describe wave shapes (P, QRS, T, ST-segment deviations) and AV conduction characteristics relevant to the disease.
        \end{itemize}
        \item Base all explanations solely on the provided information. \textbf{Do not hallucinate waveform features that cannot be inferred or documented from these inputs.}
    \end{itemize}

\item The output must be strictly in the form of JSON:
\begin{verbatim}
{
    "question": "...",
    "answer": "..."
}
\end{verbatim}
\end{enumerate}

\vspace{5mm}
\end{tcolorbox}
\end{adjustbox}
\caption{Prompt for QA pairs generation in \textit{Evidence Reasoning} tasks of \textit{Open-QA}.}
\label{fig:prompt_gen}
\end{figure*}

\noindent \textbf{Report Evaluation.} Figure~\ref{fig:prompt_eval} presents a representative prompt design for our LLM-as-a-judge evaluation system, specifically showcasing the \textit{General Summary} sub-task of \textit{Report Generation}. Acting as a strict clinical expert, the prompt evaluates the generated output against the ground truth reference across 17 granular criteria. The specific fine-grained criteria evaluated by the LLM judge are detailed previously in Table~\ref{tab:metric_stat} for \textit{Statistical Overview} and Table~\ref{tab:metric_general} for \textit{General Summary}. The prompt is explicitly designed to severely deduct points for fabricated information, missing values, or hallucinated clinical diagnoses.

\definecolor{titlebar}{RGB}{200,200,200}
\definecolor{boxbg}{RGB}{255,255,255}

\begin{figure*}[ht]
\centering
\begin{adjustbox}{width=0.95\textwidth}
\begin{tcolorbox}[
enhanced,
colback=boxbg,
colframe=black,
arc=0pt,
outer arc=0pt,
boxrule=1pt,
toprule=1.5pt,
bottomrule=1pt,
leftrule=1pt,
rightrule=1pt,
titlerule=0pt,
title={\color{black}\large\textbf{Report Evaluation Prompt}},
fonttitle=\bfseries,
attach boxed title to top left={xshift=5mm, yshift=-3mm},
boxed title style={
colback=titlebar,
colframe=titlebar,
arc=0pt,
outer arc=0pt,
boxrule=0pt,
toprule=0pt,
bottomrule=0pt,
leftrule=0pt,
rightrule=0pt,
}
]

\vspace{5mm}
% \textbf{System Prompt:} 

You are a strict and professional cardiac expert grading an LLM-generated Holter ECG report against a ground truth reference. Actively look for missing numbers, wrong values, fabricated information, and formatting issues. Deduct points severely for any discrepancy. \textbf{Do not default to 100.}

\vspace{5mm}

% \textbf{Instruction:}

% \begin{itemize}

\textbf{Reference Report:} \texttt{\{\{REFERENCE\_REPORT\}\}}\

\textbf{Generated Report:} \texttt{\{\{GENERATED\_REPORT\}\}}\

\vspace{5mm}

\textbf{Evaluation Criteria}
\begin{enumerate}
\item Accurate extraction of global metrics
\item Accurate reporting of HR extremes
\item Correct extraction of tachycardia and bradycardia burdens
\item Correct classification of the baseline rhythm (e.g., AFib)
\item Accurate total count and burden of PACs/PVCs
\item Precise breakdown of ectopic patterns (e.g., isolated, paired, runs)
\item Complete inclusion of severe events (e.g., VT runs, VF, asystole)
\item Correct interpretation of conduction blocks or pacing signals
\item Correct reporting of ST-segment and T-wave changes
\item Strict consistency with clinical facts without hallucinated findings
\item Objective description without exaggeration or understatement
\item Clinical diagnoses strictly grounded in supporting statistical data
\item Coherent narrative structure without fragmented data enumeration
\item Structured clinical hierarchy, prioritizing major diagnoses over secondary findings
\item Precise application of medical terminology
\item Consistent numeric formatting and standardized unit application (e.g., ``bpm'' for rate, ``\%'' for burden)
\item Inclusion of appropriate clinical caveats
\addtocounter{enumi}{13}
\end{enumerate}

\vspace{5mm}

\textbf{Requirements}

\begin{enumerate}

\item Score each item from 0 to 100 based on the degree of compliance with the specific criterion.

\begin{itemize}
\item \textbf{Perfect (90--100):} Perfect or near-perfect compliance with this specific criterion.
\item \textbf{Substantial (70--89):} Substantial compliance, with minor flaws, slight inaccuracies, or trivial omissions.
\item \textbf{Partial (40--69):} Captures the relevant clinical concept but applies an incorrect metric type or statistical aggregation.
\item \textbf{Poor (10--39):} Barely related or severely flawed, but not completely blank.
\item \textbf{Failure (0--9):} Complete failure; the concept is entirely missing or severely hallucinated.
\end{itemize}

\item If a specific clinical finding (e.g., ectopic patterns, severe events, ST-segment changes) is not mentioned in the reference report, it means the patient does not have it.

\begin{itemize}
\item If the generated report correctly omits it as well, this is a perfect match for that criterion. \textbf{Do not deduct points for missing a condition that isn't in the reference.}
\item However, if the reference report does not have it, but the generated report fabricates or hallucinates it, you must severely deduct points.
\end{itemize}

\item The output must be a valid JSON object where keys are the item numbers (``1'' to ``17'') and values are purely numerical scores from 0 to 100. \textbf{Do not include any other text.} For example:

\begin{verbatim}
{
    "1": 95, "2": 80, ..., "17": 100
}
\end{verbatim}

\end{enumerate}

% \end{itemize}

\vspace{5mm}
\end{tcolorbox}
\end{adjustbox}
\caption{Prompt for report evaluation in \textit{General Summary} tasks of \textit{Report Generation}.}
\label{fig:prompt_eval}
\end{figure*}

\section{Supplemental Experimental Results}
\label{app:result}

\subsection{Supplemental Generation Metrics}
\label{app:bleu}

While the main manuscript primarily focuses on clinical accuracy, specialized semantic metrics like \textit{F1-Bio} and automated LLM judge \textit{Score\textsuperscript{GPT}} to assess reasoning capabilities, standard \textit{BLEU}~\cite{papineni2002bleu} n-gram overlap metrics provide a useful supplementary perspective on textual generation quality. 

Table~\ref{tab:base_open_bleu} presents \textit{BLEU-1} and \textit{BLEU-4} scores achieved by the evaluated baseline and fine-tuned models on \textit{Open-QA} tasks, specifically evaluating \textit{Event Timing}, \textit{Diagnosis}, and \textit{Evidence Reasoning} sub-tasks.

\begin{table*}[ht]
\centering
\caption{Supplementary performance comparison of evaluated models on \textit{Open-QA} tasks from Holtercare-Bench, evaluated by additional \textit{BLEU} n-gram overlap metrics.}
\label{tab:base_open_bleu}
% \begin{adjustbox}{width=1.5\linewidth,center}
\begin{tabular}{lccccccc}
\toprule
\multirow{2}{*}{\textbf{Model}} & \multirow{2}{*}{\textbf{Modality}} & \multicolumn{2}{c}{\textbf{Event Timing}} & \multicolumn{2}{c}{\textbf{Diagnosis}} & \multicolumn{2}{c}{\textbf{Evidence Reasoning}} \\
\cmidrule(lr){3-4}
\cmidrule(lr){5-6}
\cmidrule(lr){7-8}
& & \textbf{BLEU-1} $\uparrow$ & \textbf{BLEU-4} $\uparrow$ & \textbf{BLEU-1} $\uparrow$ & \textbf{BLEU-4} $\uparrow$ & \textbf{BLEU-1} $\uparrow$ & \textbf{BLEU-4} $\uparrow$ \\

\hline
\rowcolor{gray!10}
\multicolumn{8}{c}{\textbf{\textit{Generalist Models}}} \\
\hline
\rowcolor{cyan!3}
GPT-5-mini & \faFont & 37.50 & 10.83 & 8.70 & 1.96 & 96.67 & 38.48 \\
\rowcolor{cyan!3}
Claude-4.5-Haiku & \faFont & 11.68 & 1.25 & 6.98 & 0.55 & 57.53 & 8.97 \\
\rowcolor{cyan!3}
Phi-4-mini-3.8B & \faFont & 53.57 & 13.22 & \underline{53.85} & \underline{15.73} & 96.49 & 55.56 \\
\rowcolor{orange!20}
\textbf{Phi-4-mini-3.8B\textsuperscript{FT}} & \faFont & \textbf{95.00} & \textbf{77.39} & \textbf{98.46} & \textbf{87.05} & \textbf{93.70} & \textbf{72.64} \\
\rowcolor{lime!3}
InternVL-3.5-8B & \faFilm & 41.46 & 8.65 & 35.29 & 6.02 & 96.30 & 30.39 \\
\rowcolor{lime!3}
MiniCPM-V4.5-8B & \faFilm & 16.33 & 2.06 & 15.79 & 1.25 & 91.49 & 35.50 \\
\rowcolor{lime!3}
Gemini-3.0-Flash & \faFilm & \underline{66.67} & 22.93 & 17.65 & 2.87 & 96.88 & 45.06 \\
\rowcolor{lime!3}
Qwen3-VL-8B & \faFilm & 56.67 & 12.68 & 50.00 & 0.00 & 95.35 & 52.34 \\
\rowcolor{orange!20}
\textbf{Qwen3-VL-8B\textsuperscript{FT}} & \faFilm & 60.95 & \underline{31.91} & 30.77 & 12.36 & 97.14 & 50.30 \\

\hline
\rowcolor{gray!10}
\multicolumn{8}{c}{\textbf{\textit{Medical Models}}} \\
\hline
\rowcolor{cyan!3}
LLaVA-Med-V1.5-7B & \faFont & 37.25 & 7.50 & 37.50 & 5.45 & 86.05 & 42.02 \\
\rowcolor{cyan!3}
MedGemma-1.5-4B-IT & \faFont & 22.22 & 2.90 & 16.00 & 2.13 & 97.62 & \underline{65.55} \\
\rowcolor{cyan!3}
HealthGPT-M3-3.8B & \faFont & 34.21 & 6.19 & 12.36 & 1.24 & 90.24 & 43.46 \\
\rowcolor{lime!3}
Lingshu-7B & \faFilm & 33.33 & 5.61 & 25.00 & 3.98 & 97.67 & 47.90 \\
\rowcolor{lime!3}
MedVLM-R1-2B & \faFilm & 40.54 & 6.60 & 33.33 & 4.50 & 85.96 & 23.23 \\
\rowcolor{lime!3}
HuatuoGPT-Vision-7B & \faFilm & 53.33 & 21.02 & 33.33 & 8.05 & \underline{98.32} & 8.16 \\

\bottomrule
\end{tabular}
% \end{adjustbox}
\end{table*}

Table ~\ref{tab:base_report_bleu} details the corresponding \textit{BLEU-1} and \textit{BLEU-4} metrics for \textit{Report Generation} tasks, covering both \textit{Statistical Overview} and \textit{General Summary}. These supplemental metrics highlight the lexical and structural alignment between the models' generated responses and the expert-crafted ground truth.

\begin{table*}[ht]
\centering
\caption{Supplementary performance comparison of evaluated models on \textit{Report Generation} tasks from Holtercare-Bench, evaluated by additional \textit{BLEU} n-gram overlap metrics.}
\label{tab:base_report_bleu}
% \begin{adjustbox}{width=1.3\linewidth,center}
\begin{tabular}{lccccc}
\toprule
\multirow{2}{*}{\textbf{Model}} & \multirow{2}{*}{\textbf{Modality}} & \multicolumn{2}{c}{\textbf{Statistical Overview}} & \multicolumn{2}{c}{\textbf{General Summary}} \\
\cmidrule(lr){3-4}
\cmidrule(lr){5-6}
& & \textbf{BLEU-1} $\uparrow$ & \textbf{BLEU-4} $\uparrow$ & \textbf{BLEU-1} $\uparrow$ & \textbf{BLEU-4} $\uparrow$ \\

\hline
\rowcolor{gray!10}
\multicolumn{6}{c}{\textbf{\textit{Generalist Models}}} \\
\hline
\rowcolor{cyan!3}
GPT-5-mini & \faFont & 44.64 & 3.65 & 60.47 & 6.95 \\
\rowcolor{cyan!3}
Claude-4.5-Haiku & \faFont & 15.92 & 1.15 & 27.60 & 1.00 \\
\rowcolor{cyan!3}
Phi-4-mini-3.8B & \faFont & 56.76 & 10.99 & 52.43 & 6.03 \\
\rowcolor{orange!20}
\textbf{Phi-4-mini-3.8B\textsuperscript{FT}} & \faFont & \textbf{97.67} & \textbf{68.03} & \textbf{94.83} & \textbf{71.94} \\
\rowcolor{lime!3}
InternVL-3.5-8B & \faFilm & \underline{81.63} & \underline{42.20} & \underline{71.43} & \underline{14.88} \\
\rowcolor{lime!3}
MiniCPM-V4.5-8B & \faFilm & 42.55 & 6.95 & 61.96 & 12.88 \\
\rowcolor{lime!3}
Gemini-3.0-Flash & \faFilm & 51.11 & 5.48 & 67.07 & 9.86 \\
\rowcolor{lime!3}
Qwen3-VL-8B & \faFilm & 34.75 & 8.53 & 41.13 & 3.22 \\
\rowcolor{orange!20}
\textbf{Qwen3-VL-8B\textsuperscript{FT}} & \faFilm & 47.31 & 4.88 & 50.56 & 4.73 \\

\hline
\rowcolor{gray!10}
\multicolumn{6}{c}{\textbf{\textit{Medical Models}}} \\
\hline
\rowcolor{cyan!3}
LLaVA-Med-V1.5-7B & \faFont & 45.76 & 3.53 & 58.73 & 8.09 \\
\rowcolor{cyan!3}
MedGemma-1.5-4B-IT & \faFont & 59.62 & 9.42 & 46.81 & 3.94 \\
\rowcolor{cyan!3}
HealthGPT-M3-3.8B & \faFont & 10.79 & 1.47 & 52.63 & 10.27 \\
\rowcolor{lime!3}
Lingshu-7B & \faFilm & 38.82 & 7.56 & 59.82 & 10.40 \\
\rowcolor{lime!3}
MedVLM-R1-2B & \faFilm & 68.29 & 18.42 & 60.00 & 7.18 \\
\rowcolor{lime!3}
HuatuoGPT-Vision-7B & \faFilm & 37.63 & 4.73 & 52.14 & 7.05 \\

\bottomrule
\end{tabular}
% \end{adjustbox}
\end{table*}

\subsection{Modality Alignment and Representation Analysis}
\label{app:text}

To accommodate the diverse architectural constraints of contemporary MLLMs, our data engine, HolterAgent, constructs a format compatibility pipeline that transforms a single, unified \texttt{.edf} signal source into three distinct representation formats: raw signal, video stream, and clinical text. This design is fundamentally intended to maximize MLLM compatibility rather than to introduce three independent data sources. Specifically, the video modality is prioritized for its ability to preserve the continuous temporal dynamics and morphological evolution of streaming electrophysiological data. Conversely, the text modality serves as a robust fallback for models lacking native video processing capabilities, representing the signal through discretized numerical sequences.

To empirically verify that this multi-representation design does not introduce modality-specific artifacts or bias the evaluation, we conduct a comprehensive comparative analysis. Table~\ref{tab:base_closed_text} presents the \textit{Closed-QA} performance of all video-capable models when evaluated exclusively under the text modality. The results demonstrate that neither modality consistently dominates across all models and tasks. For instance, while certain models exhibit marginal improvements in specific tasks under the text modality, others show a clear preference for video inputs. This non-uniform performance distribution confirms that the observed performance gaps among different MLLMs reflect their intrinsic architectural capabilities and inductive biases in processing long-term physiological sequences, rather than artifacts induced by our data formatting pipeline.

\begin{table*}[ht]
\centering
\caption{Supplementary performance comparison of video-capable models evaluated under the text modality on \textit{Closed-QA} tasks from Holtercare-Bench, evaluated by accuracy.}
\label{tab:base_closed_text}
\begin{tabular}{lcccccc}
\toprule
\textbf{Model} & \textbf{Modality} & \textbf{Presence} & \makecell{\textbf{Event}\\ \textbf{Counting}} & \makecell{\textbf{Event}\\ \textbf{Timing}} & \makecell{\textbf{HR Extremum}\\ \textbf{Timing}} & \textbf{Diagnosis} \\

\hline
\rowcolor{gray!10}
\multicolumn{7}{c}{\textbf{\textit{Generalist Models}}} \\
\hline
\rowcolor{cyan!3}
InternVL-3.5-8B & \faFont & 58.01 & 29.66 & 47.26 & 31.85 & 39.37 \\
\rowcolor{cyan!3}
MiniCPM-V4.5-8B & \faFont & 54.90 & 51.38 & 28.05 & 35.99 & 26.38 \\
\rowcolor{cyan!3}
Gemini-3.0-Flash & \faFont & 67.32 & 14.98 & 62.50 & 53.18 & 41.73 \\
\rowcolor{cyan!3}
Qwen3-VL-8B & \faFont & 60.95 & 31.80 & 28.66 & 34.71 & 32.28 \\
\rowcolor{orange!20}
\textbf{Qwen3-VL-8B\textsuperscript{FT}} & \faFont & 95.10 & 83.79 & 97.23 & 71.97 & 94.88 \\

\hline
\rowcolor{gray!10}
\multicolumn{7}{c}{\textbf{\textit{Medical Models}}} \\
\hline
\rowcolor{cyan!3}
Lingshu-7B & \faFont & 47.55 & 35.17 & 24.70 & 29.94 & 25.98 \\
\rowcolor{cyan!3}
MedVLM-R1-2B & \faFont & 40.20 & 41.28 & 26.52 & 32.17 & 13.39 \\
\rowcolor{cyan!3}
HuatuoGPT-Vision-7B & \faFont & 52.29 & 25.38 & 26.22 & 32.17 & 23.23 \\
\bottomrule
\end{tabular}
\end{table*}

\subsection{Statistical Significance Analysis}
\label{app:mcnemar}

To confirm whether the substantial performance improvements observed after fine-tuning are statistically robust and not attributable to random sampling variance or favorable test-set splits, we perform McNemar's tests on all five \textit{Closed-QA} tasks for both fine-tuned representative models, Phi-4-mini-3.8B~\cite{abdin2024phi} and Qwen3-VL-8B~\cite{bai2025qwen3}. As detailed in Table~\ref{tab:base_mcnemar}, all pairwise comparisons between the zero-shot baselines and fine-tuned models yield $p$-values $\ll 0.001$. These statistical significances across all tasks and both models provide evidence that the fine-tuning process induces genuine capability acquisition and knowledge internalization in long-context ECG reasoning, rather than mere fluctuations in sampling variance.

\begin{table*}[ht]
\centering
\caption{Statistical significance analysis of improvements from zero-shot baselines to fine-tuned models on \textit{Closed-QA} tasks from Holtercare-Bench. The reported $p$-values are derived from McNemar tests.}
\label{tab:base_mcnemar}
\begin{tabular}{lcccccc}
\toprule
\textbf{Model} & \textbf{Modality} & \textbf{Presence} & \makecell{\textbf{Event}\\ \textbf{Counting}} & \makecell{\textbf{Event}\\ \textbf{Timing}} & \makecell{\textbf{HR Extremum}\\ \textbf{Timing}} & \textbf{Diagnosis} \\
\midrule
\rowcolor{cyan!10}
Phi-4-mini-3.8B & \faFont & $1.95 \times 10^{-5}$ & $3.99 \times 10^{-16}$ & $6.33 \times 10^{-20}$ & $3.09 \times 10^{-8}$ & $1.76 \times 10^{-11}$ \\
\rowcolor{cyan!10}
Qwen3-VL-8B & \faFont & $6.17 \times 10^{-41}$ & $5.26 \times 10^{-32}$ & $2.18 \times 10^{-52}$ & $6.57 \times 10^{-18}$ & $3.55 \times 10^{-35}$ \\
\rowcolor{lime!10}
Qwen3-VL-8B & \faFilm & $7.08 \times 10^{-44}$ & $3.14 \times 10^{-36}$ & $6.14 \times 10^{-47}$ & $1.08 \times 10^{-17}$ & $1.30 \times 10^{-28}$ \\
\bottomrule
\end{tabular}
\end{table*}

\end{document}